\documentclass[english,final]{IEEEtran} 

\makeatletter

\usepackage[mathscr]{eucal}
\usepackage{epsfig,epsf,psfrag}
\usepackage{amssymb,amsmath,amsfonts,latexsym}
\usepackage{amsmath,graphicx,bm,xcolor,url}
\usepackage[caption=false]{subfig} 
\usepackage{fixltx2e}
\usepackage{array}
\usepackage{verbatim}
\usepackage{bm}
\usepackage{algpseudocode}
\usepackage{algorithm}
\usepackage{verbatim}
\usepackage{textcomp}
\usepackage{mathrsfs}
\usepackage{epstopdf}
\usepackage{relsize}
\usepackage{cleveref} 
\usepackage{subfig}
 \usepackage{amsthm}
 \usepackage{dsfont}

\usepackage{tikz}
\usepackage{pgfplots}
\usepgfplotslibrary{groupplots} 
\usetikzlibrary{pgfplots.groupplots}
\usetikzlibrary{matrix,arrows,decorations.pathmorphing}
\usetikzlibrary{shapes,snakes}
\usepackage{pgfplots,pgfplotstable}
\usepgfplotslibrary{fillbetween}
\pgfplotsset{compat=1.10}

\catcode`~=11 \def\UrlSpecials{\do\~{\kern -.15em\lower .7ex\hbox{~}\kern .04em}} \catcode`~=13 

\allowdisplaybreaks[3]

\newcommand{\norm}[1]{\left\Vert#1\right\Vert}

\newcommand{\SREC}{\mathrm{S}\textrm{-}\mathrm{REC}}

\newcommand{\boldeta}{\boldsymbol{\eta}}

\newcommand{\calA}{\mathcal{A}}
\newcommand{\calB}{\mathcal{B}}

\newcommand{\calN}{\mathcal{N}}

\newcommand{\calS}{\mathcal{S}}

\newcommand{\calX}{\mathcal{X}}

\newcommand{\bA}{\mathbf{A}}

\newcommand{\bI}{\mathbf{I}}

\newcommand{\bM}{\mathbf{M}}

\newcommand{\bQ}{\mathbf{Q}}

\newcommand{\bt}{\mathbf{t}}

\newcommand{\bu}{\mathbf{u}}

\newcommand{\bv}{\mathbf{v}}

\newcommand{\bw}{\mathbf{w}}
\newcommand{\bW}{\mathbf{W}}
\newcommand{\bx}{\mathbf{x}}

\newcommand{\by}{\mathbf{y}}

\newcommand{\bz}{\mathbf{z}}

\newcommand{\rmF}{\mathrm{F}}

\newcommand{\bbE}{\mathbb{E}}

\newcommand{\bbR}{\mathbb{R}}

\DeclareMathAlphabet{\mathbsf}{OT1}{cmss}{bx}{n}
\DeclareMathAlphabet{\mathssf}{OT1}{cmss}{m}{sl}

\DeclareSymbolFont{bsfletters}{OT1}{cmss}{bx}{n}  
\DeclareSymbolFont{ssfletters}{OT1}{cmss}{m}{n}
\DeclareMathSymbol{\bsfGamma}{0}{bsfletters}{'000}
\DeclareMathSymbol{\ssfGamma}{0}{ssfletters}{'000}
\DeclareMathSymbol{\bsfDelta}{0}{bsfletters}{'001}
\DeclareMathSymbol{\ssfDelta}{0}{ssfletters}{'001}
\DeclareMathSymbol{\bsfTheta}{0}{bsfletters}{'002}
\DeclareMathSymbol{\ssfTheta}{0}{ssfletters}{'002}
\DeclareMathSymbol{\bsfLambda}{0}{bsfletters}{'003}
\DeclareMathSymbol{\ssfLambda}{0}{ssfletters}{'003}
\DeclareMathSymbol{\bsfXi}{0}{bsfletters}{'004}
\DeclareMathSymbol{\ssfXi}{0}{ssfletters}{'004}
\DeclareMathSymbol{\bsfPi}{0}{bsfletters}{'005}
\DeclareMathSymbol{\ssfPi}{0}{ssfletters}{'005}
\DeclareMathSymbol{\bsfSigma}{0}{bsfletters}{'006}
\DeclareMathSymbol{\ssfSigma}{0}{ssfletters}{'006}
\DeclareMathSymbol{\bsfUpsilon}{0}{bsfletters}{'007}
\DeclareMathSymbol{\ssfUpsilon}{0}{ssfletters}{'007}
\DeclareMathSymbol{\bsfPhi}{0}{bsfletters}{'010}
\DeclareMathSymbol{\ssfPhi}{0}{ssfletters}{'010}
\DeclareMathSymbol{\bsfPsi}{0}{bsfletters}{'011}
\DeclareMathSymbol{\ssfPsi}{0}{ssfletters}{'011}
\DeclareMathSymbol{\bsfOmega}{0}{bsfletters}{'012}
\DeclareMathSymbol{\ssfOmega}{0}{ssfletters}{'012}

\DeclareMathOperator*{\argmin}{arg\,min}

\DeclareMathOperator{\vect}{vec}

\newcommand{\bzero}{\mathbf{0}}
\newcommand{\bone}{\mathds{1}}

\theoremstyle{plain}
\newtheorem{theorem}{Theorem} 

\newtheorem{proposition}{Proposition}

\newtheorem{definition}{Definition}

\newcommand{\qednew}{\nobreak \ifvmode \relax \else
      \ifdim\lastskip<1.5em \hskip-\lastskip
      \hskip1.5em plus0em minus0.5em \fi \nobreak
      \vrule height0.75em width0.5em depth0.25em\fi}

\DeclareMathOperator{\relu}{\mathrm{relu}}
\DeclareMathOperator{\sspan}{\mathrm{span}}
\newcommand{\R}{\mathbb{R}}
\newcommand{\eps}{\epsilon}

\newcommand{\zo}{\bz^*}

\newcommand{\zopt}{\bz}
\newcommand{\xopt}{\bx}
\newcommand{\vzoptzo}{\bv_{\zopt, \zo}}

\newcommand{\xtrue}{\bx^*}

\floatstyle{ruled}
\newfloat{algorithm}{tbp}{loa}
\providecommand{\algorithmname}{Algorithm}
\floatname{algorithm}{\protect\algorithmname}

\makeatletter
\newcommand{\manuallabel}[2]{\def\@currentlabel{#2}\label{#1}}
\makeatother

\begin{document} 
    
    \title{Theoretical Perspectives on Deep \\ Learning Methods in Inverse Problems}
    \author{Jonathan Scarlett, Reinhard Heckel, Miguel R.~D.~Rodrigues, Paul Hand, and Yonina C.~Eldar \thanks{J.~Scarlett is with the Department of Computer Science, Department of Mathematics, and Institute of Data Science, National University of Singapore, Singapore. (e-mail: scarlett@comp.nus.edu.sg)}
    \thanks{R.~Heckel is with the Department of Electrical and Computer Engineering, Technical University of Munich, Germany. (e-mail: reinhard.heckel@tum.de)}
    \thanks{M.~R.~D.~Rodrigues is with the Department of Electronic and Electrical Engineering, University College London, UK. (e-mail: m.rodrigues@ucl.ac.uk)}
    \thanks{P.~Hand is with the College of Science and the Khoury College of Computer Sciences, Northeastern University, USA. (e-mail: p.hand@northeastern.edu)}
    \thanks{Y.~C.~Eldar is with the Department of Computer Science and Applied Mathematics, Weizmann Institute of Science, Israel. (e-mail: yonina.eldar@weizmann.ac.il)}}
    \maketitle
    
    \begin{abstract}
        In recent years, there have been significant advances in the use of deep learning methods in inverse problems such as denoising, compressive sensing, inpainting, and super-resolution.  While this line of works has predominantly been driven by practical algorithms and experiments, it has also given rise to a variety of intriguing theoretical problems.  In this paper, we survey some of the prominent theoretical developments in this line of works, focusing in particular on generative priors, untrained neural network priors, and unfolding algorithms.  In addition to summarizing existing results in these topics, we highlight several ongoing challenges and open problems.
    \end{abstract}

        \section{Introduction} \label{sec:intro}
        
    The study of inverse problems spans several research communities, covering problems such as inpainting, denoising, super-resolution, medical imaging, and more.  Over the years, research on inverse problems has seen a series of paradigm shifts and new perspectives; for instance, the incorporation of low-dimensional structure such as sparsity led to extensive research on \emph{compressive sensing} \cite{foucart2013mathematical,eldar2015sampling,DDEK11}.
    
    The most prominent new trend in inverse problems is the incorporation of \emph{deep learning} methods, which have been utilized for signal modeling, decoder design, measurement design, and more. These methods frequently attain state-of-the-art performance in domains such as imaging, signal processing, and communications.  While research in this direction has predominantly been practically-oriented and relied on experiments for evaluation, it has also given rise to a wide variety of interesting theoretical developments and challenges.  In this paper, we provide an introductory overview of theoretical frameworks and results relating to deep learning methods in inverse problems, and highlight their strengths, limitations, and directions for further research.
    
    \subsection{Background: Inverse Problems}
    
    The goal of an inverse problem is to recover (either exactly or approximately) an unknown signal $\xtrue \in \bbR^n$ from a set of measurements $\by \in \bbR^m$ (often referred to as observations),\footnote{The reals can also be replaced by the complex numbers or other mathematical types, depending on the application.} which are related via a \emph{measurement model} $\calA$ (often referred to as the \emph{forward model}):
    \begin{equation}
        \by = \calA(\xtrue) + \boldeta, \label{eq:y_general}
    \end{equation}
    where $\boldeta$ represents possible additive noise.\footnote{More generally, we could write $\calA(\xtrue,\boldeta)$ to model noise that need not be additive.}  The measurement model $\calA$ may be known, unknown, or partially known. 
    
    An important special case is the class of \emph{linear models}, in which $\calA$ is a linear operation:
    \begin{equation}
        \by = \bA \xtrue + \boldeta \label{eq:y_linear}
    \end{equation}
    for some \emph{measurement matrix} $\bA \in \bbR^{m \times n}$.  We focus on the case that $\bA$ is known (unless stated otherwise), and the goal is to design an algorithm 
    that recovers $\xtrue$ from $(\bA,\by)$.
    
    Linear models already capture numerous important problems, including denoising, inpainting, deblurring, and super-resolution.  Among these, we highlight the seemingly simplest problem of denoising, in which $\bA$ is the identity matrix:
    \begin{equation}
        \by = \xtrue + \boldeta.
    \end{equation}
    While this problem may appear limited in scope compared to general linear or non-linear measurement models, it turns out that effective solutions to the denoising problem 
    can be used as a powerful building block to solve more general inverse problems via plug-and-play methods \cite{venkatakrishnan2013plug,metzler2016denoising,romano2017little}. 
    
    A crucial component of inverse problems and their associated algorithms/theory is the assumed prior knowledge on the underlying signal $\xtrue$.  Such prior knowledge typically amounts to an assumption that $\xtrue$ lies in or near some restricted set $\calX$, which may be intrinsically low-dimensional despite $\bbR^n$ being a high-dimensional space.  A ubiquitous example is the set of sparse signals:
    \begin{equation}
        \calX_s = \big\{ \xopt \in \bbR^n \,:\, \|\xopt\|_0 \le s \big\}, \label{eq:X_sparse}
    \end{equation}
    where $\|\xopt\|_0$ denotes the number of non-zero entries in $\xopt$, and $s$ is a suitably-chosen sparsity level, typically with $s \ll n$.  Related notions include structured sparsity \cite{baraniuk2010model,EM09a}, low-rankness \cite{recht2010guaranteed}, and manifold structure \cite{baraniuk2009random,hegde2012signal}.
    
    \subsection{Deep Learning Methods for Solving Inverse Problems} \label{sec:dl_inverse}
    
    Advances in neural networks and deep learning have reshaped the field of machine learning, and are increasingly impacting other domains throughout academia and industry.  
    As hinted above, inverse problems are no exception to this trend.  Previous surveys on deep learning methods in inverse problems can be found in \cite{mccann2017convolutional,ongie2020deep}, and the key distinction of our survey is our focus on mathematical theory.  The reader is assumed to be familiar with basic neural network concepts such as depth, width, training, empirical risk minimization, gradient descent, generalization, convolutional neural networks, and recurrent neural networks; an introduction to these concepts can be found in \cite{zhang2022dive}, among many others.
    
    There are many different ways in which deep learning can play a role in designing methods for inverse problems.  We will focus on the following three themes in this survey:

     \subsubsection{Generative priors} 
     One of the tremendous successes of deep learning has been \emph{deep generative modeling}, in which a neural network is trained on a large data set of signals/images, and the resulting network $G \colon \bbR^k \to \bbR^n$ (typically with $k \ll n$) serves as a model for the underlying class of signals, i.e., for each input $\bz \in \bbR^k$, the output $G(\bz)$ corresponds to some signal (or image in vectorized form).  The network is \emph{generative} in the sense that it can generate new images different to those used for training.
    
    Building on practically-oriented works such as \cite{dong2015image,yeh2016semantic,lipton2017precise}, Bora \emph{et al.}~\cite{bora2017compressed} introduced a theoretical framework for studying generative model based priors in inverse problems.  In comparison to sparse modeling, the idea is to replace the set $\calX_s$ in \eqref{eq:X_sparse} by the set
    \begin{equation}
        \calX_G = \mathrm{Range}(G).
    \end{equation}
    By doing so, the prior knowledge can be much more specifically geared to the task at hand.  For instance, while a sparse prior in a suitably-chosen basis could model nearly all natural images, a generative prior could specifically target a particular type of image (e.g., brain scans in medical imaging), thus providing a much more precise form of prior information, and leading to improved reconstruction accuracy and/or fewer required measurements.  We survey several relevant theoretical results in Section \ref{sec:generative_stat}.
    
    \subsubsection{Untrained neural network priors} 
    It has recently been observed that even neural networks with \emph{no prior training} can serve as excellent priors for inverse problems \cite{ulyanov_DeepImagePrior_2018,heckel_DeepDecoderConcise_2019}.  In this approach, the prior information is implicitly encoded in the neural network architecture, and decoding is done by tuning the weights to produce a single image that fits the measurements well.  
    
    Despite using neural networks, these methods are perhaps more closely related to sparse priors, in the sense that the priors are ``broad'' (e.g., capturing general natural images) and are not targeted at specific data sets.  On the other hand, their empirical performance often significantly improves on that of sparsity-based methods.  We survey several relevant theoretical developments in Section \ref{sec:untrained}.

    \subsubsection{Unfolding methods} 
    Another component of inverse problems amenable to deep learning methods is the design of the decoder, e.g., the algorithm for reconstructing $\bx^*$ from $(\bA,\by)$ in the case of linear measurements.   
    A variety of deep learning approaches have been devised for this task, consisting of trainable components that are optimized for the task at hand, e.g., see \cite{ongie2020deep,mccann2017convolutional,SEMBAI,shlezinger2022model} for recent surveys.
    
    
    In Section \ref{sec:unfolding}, we consider sparse signal priors and survey the prominent approach of \emph{algorithm unfolding} \cite{gregor2010learning,monga2021algorithm}, which frequently provides state-of-the-art practical performance.  Briefly, the idea is to select a (recurrent) neural network structure that directly matches a classical iterative algorithm, but to replace the fixed weights of that algorithm with learnable weights.  A detailed survey of algorithm unfolding techniques can be found in \cite{monga2021algorithm}, and our survey is again distinguished by the focus on theory.
    
    \medskip
    These three topics are by no means exhaustive; for instance, there are many deep learning based decoders beyond unfolding methods \cite{ongie2020deep,mccann2017convolutional,SEMBAI,shlezinger2022model} (as mentioned above), and there are other aspects of inverse problems that also admit deep learning methods, such as designing the measurement matrix \cite{wu2019learning,learn2sample}.  In Section \ref{sec:other}, we will briefly discuss some further relevant topics beyond the three that we focus on.

\subsection{Theoretical Guarantees for Sparse Recovery} \label{sec:sparse_rec}

To set the stage for the results that we overview in this paper, it is useful to summarize some of the related results in the literature on sparse recovery.  For concreteness, we focus on linear models of the form \eqref{eq:y_linear}, and signals that are exactly or approximate sparse according to \eqref{eq:X_sparse}, though many results are known beyond this setting (e.g., see \cite{foucart2013mathematical}).  Among the wide range of concepts and results in the literature, we focus on a small sample that are particularly relevant to this survey, and for which we consider closely-related notions for deep learning methods throughout Sections \ref{sec:generative_stat}--\ref{sec:unfolding}.

\textbf{Recovery guarantees.} Theoretical results on sparse recovery can differ considerably depending on the presence/absence of noise, whether the signal is exactly or approximately sparse, and the desired recovery guarantee.  Particularly relevant to this survey is the \emph{$\ell_2/\ell_2$ for-each guarantee}, which states that there exists a randomized measurement matrix $\bA$ such that given $\by=\bA\xtrue$ (and $\bA$), the decoder outputs some $\hat{\bx}$ satisfying the following with high probability:
\begin{equation}
    \|\hat{\bx}-\xtrue\|_2 \le C \min_{\xopt \in \calX_s} \|\xopt-\xtrue\|_2 \label{eq:l2l2}
\end{equation}
for some $C > 1$.  That is, the estimation error is within a constant factor of the best possible sparse approximation.  
This guarantee can be achieved with constant probability and $m = O\big( s \log \frac{n}{s} \big)$ \cite{cohen2009compressed}, or more generally, with probability $1-\rho$ and $m = O\big( s \log \frac{n}{s} + \log\frac{1}{\rho} \big)$ \cite{gilbert2013ell}.

To highlight the impact of the recovery criteria, we note that deterministically attaining \eqref{eq:l2l2} (for all $\bx^*$) with fixed $\bA$ is only possible when $m = \Omega(n)$ \cite{cohen2009compressed}, though analogous guarantees are possible by using different norms on the left and right sides of \eqref{eq:l2l2}, known as $\ell_p$/$\ell_q$ guarantees (e.g., $p=q=1$).  In contrast, when $\xtrue$ is exactly sparse and the measurements are noisy (i.e., $\by = \bA\xtrue + \boldeta$), the preceding difficulty is alleviated, and one can attain a deterministic guarantee of the form
\begin{equation}
    \|\hat{\bx} - \xtrue\|_2 \le C \|\boldeta\|_2 \label{eq:deterministic}
\end{equation}
for some constant $C$, with $m = O\big( s \log \frac{n}{s} \big)$ \cite{candes2006stable}.

Importantly, the guarantees \eqref{eq:l2l2} and \eqref{eq:deterministic} (as well as other related guarantees) with the above-mentioned bounds on $m$ are not only information-theoretically achievable, but are known to be attained by \emph{practical} decoding algorithms coupled with suitably-chosen $\bA$.  Some common choices of $\bA$ and decoding algorithms are discussed below.

\textbf{Measurement matrix design and properties.} The measurement matrix $\bA$ is often constrained by the application (e.g., subsampled Fourier matrices in medical imaging), but can sometimes be designed freely.  In theoretical studies, the most widely-considered type of measurement matrix is \emph{i.i.d.~Gaussian}, in which each entry of $\bA$ is independently drawn from $\calN(0,1)$, $\calN\big(0,\frac{1}{m}\big)$, or similar (the choice of normalization varies for convenience of the analysis).  For probabilistic guarantees such as \eqref{eq:l2l2}, such designs are often analyzed directly.  For deterministic guarantees such as \eqref{eq:deterministic} the typical approach is to (i) establish deterministic conditions on $\bA$ that suffice to obtain the desired recovery guarantee, and (ii) establish that i.i.d.~Gaussian (or other randomized) measurements satisfy those conditions with high probability.

We highlight in particular the \emph{restricted isometry property}  (RIP) \cite{candes2005decoding}: The matrix $\bA$ satisfies the RIP with parameters ($s$, $\delta_{s}$) if, for every $\bx \in \calX_{s}$, it holds that 
\begin{equation}
    (1-\delta_{s})\|\bx\|_2^2 \le \|\bA\bx\|_2^2 \le (1+\delta_{s})\|\bx\|_2^2. \label{eq:rip}
\end{equation}
Intuitively, this property states that $\bA$ is nearly orthonormal when restricted to sparse vectors.  Certain works instead only required the lower bound on $\|\bA\bx\|_2^2$ in \eqref{eq:rip}, and this variant is known as the \emph{restricted eigenvalue condition} (REC) \cite{bickel2009simultaneous}. 

\textbf{Information-theoretic lower bounds.} In the above discussion, we highlighted that various upper bounds on the number of measurements have been obtained for attaining recovery guarantees such as \eqref{eq:l2l2} and \eqref{eq:deterministic}.  These are complemented by \emph{information-theoretic lower bounds}, which state that any sparse recovery algorithm attaining a certain guarantee must have a minimum number of measurements.  Such results are crucial in certifying the degree of optimality of practical algorithms, and steering research towards cases where the greatest improvements are possible.

Lower bounds for sparse recovery have been obtained for a variety of recovery criteria (e.g., see \cite{candes2013how,ba2010lower,price2011epsilon,gilbert2013ell}), often with scaling laws that match existing upper bounds.  Among these, we highlight the fact that any algorithm attaining the $\ell_2$/$\ell_2$ guarantee in \eqref{eq:l2l2} with constant probability must have $m = \Omega\big(s \log \frac{n}{s} \big)$, thus matching the above-mentioned upper bound to within a constant factor.  A proof of this result is given in \cite{price2011epsilon}, based on a reduction to a communication problem over a Gaussian channel.

\textbf{Practical decoding techniques.} Recovery guarantees, often with a near-optimal number of measurements, have been attained for a wide range of practical decoding techniques.  For instance, the RIP and/or REC have been used as a tool for studying guarantees of convex relaxation algorithms, thresholding algorithms, and greedy algorithms (e.g., see \cite[Ch.~6]{foucart2013mathematical} and \cite{bickel2009simultaneous}).  The class of convex relaxation algorithms can roughly be viewed as trying to find $\xopt$ such that both $\|\by - \bA\xopt\|_2$ and $\|\xopt\|_0$ (the number of non-zeros in $\xopt$) are small, but to circumvent the combinatorial nature of the latter, the convex proxy $\|\bx\|_1$ is used.  A famous example is the least absolute shrinkage and selection operator (Lasso) method, in which $\hat{\bx}$ is the solution to
\begin{align}
    \min_{\xopt} \| \by - \bA\xopt \|_2^2 + \lambda  \| \xopt \|_1 \label{eq:Lasso}
\end{align}
for some regularization parameter $\lambda > 0$.  This is a convex optimization problem for which numerous solvers are available that converge to the optimal solution. 

In principle, \eqref{eq:Lasso} could be solved using off-the-shelf convex optimization solvers, but due to the ubiquity of Lasso, several special-purpose iterative algorithms have also been devised.  In Section \ref{sec:unfolding}, one such algorithm called the iterative shrinkage thresholding algorithm (ISTA) \cite{daubechies2004iterative} will play a major role.

\subsection{Overview of the Paper}

Our goal is to provide an introduction to several theoretical results on deep learning methods in inverse problems.  In addition, we seek to highlight interesting connections between these results, and to discuss ongoing challenges and open problems.  We provide intuition behind several of the associated proofs, but avoid going into significant technical detail.

The structure of the paper is as follows:
\begin{itemize}
    \item In Section \ref{sec:generative_stat}, we overview several theoretical developments concerning generative priors in inverse problems, including statistical guarantees, information-theoretic limits, and optimization guarantees.
    \item In Section \ref{sec:untrained}, we overview theoretical developments regarding neural network priors with no prior training, including provable recovery guarantees for denoising and compressive sensing.
    \item In Section \ref{sec:unfolding}, we overview theoretical developments regarding unfolding algorithms, focusing on sparse signal priors and neural network structures that are based on the classical ISTA algorithm.
    \item In Section \ref{sec:other}, we discuss other uses of deep learning in inverse problems, highlighting additional relevant existing theory, as well as scenarios where theory is currently lacking but may be of interest.  Several directions for future research are additionally mentioned throughout Sections \ref{sec:generative_stat}--\ref{sec:unfolding}.
\end{itemize}
We emphasize that our goal is not to be exhaustive or near-exhaustive in covering the existing literature.  While we seek to cover a diverse set of perspectives and results, the ones that we focus on are naturally heavily influenced by our own backgrounds and interests.

\medskip
\textbf{Notation.} We make frequent use of the standard asymptotic notation $O(\cdot)$ and $\Omega(\cdot)$ (note that $f_n = \Omega(g_n) \iff g_n = O(f_n)$).  
The ReLU function is given by $\relu(z) = \max\{0,z\}$, and is applied element-wise when applied to vectors.  Further notation will be introduced throughout the relevant sections.

    \section{Generative Priors} \label{sec:generative_stat} 

In this section, we overview a recent line of works studying theoretical guarantees for inverse problems with generative priors.  We begin by outlining the relevant background, and then state some statistical upper and lower bounds.  We then turn to guarantees for specific optimization procedures.

\subsection{Background}

As outlined in Section \ref{sec:dl_inverse}, the idea of this line of works is to replace conventional priors (e.g., sparse or low-rank models) by \emph{data-driven generative priors} that can be much more specifically targeted to the task at hand.  Given a generative network $G \colon \bbR^k \to \bbR^n$ that accurately models the signals we are interested in, it is natural to decode by outputting a signal in ${\rm Range}(G)$ that best matches the measurements in some sense (e.g., $\|\by - \bA G(\bz)\|_2$ is small).  This idea is captured by equations \eqref{eq:x_Gz}--\eqref{eq:z_argmin} and \eqref{eq:x_Gz2}--\eqref{eq:z_argmin2} to follow.

\begin{figure}
    \begin{centering}
        \includegraphics[width=0.65\columnwidth]{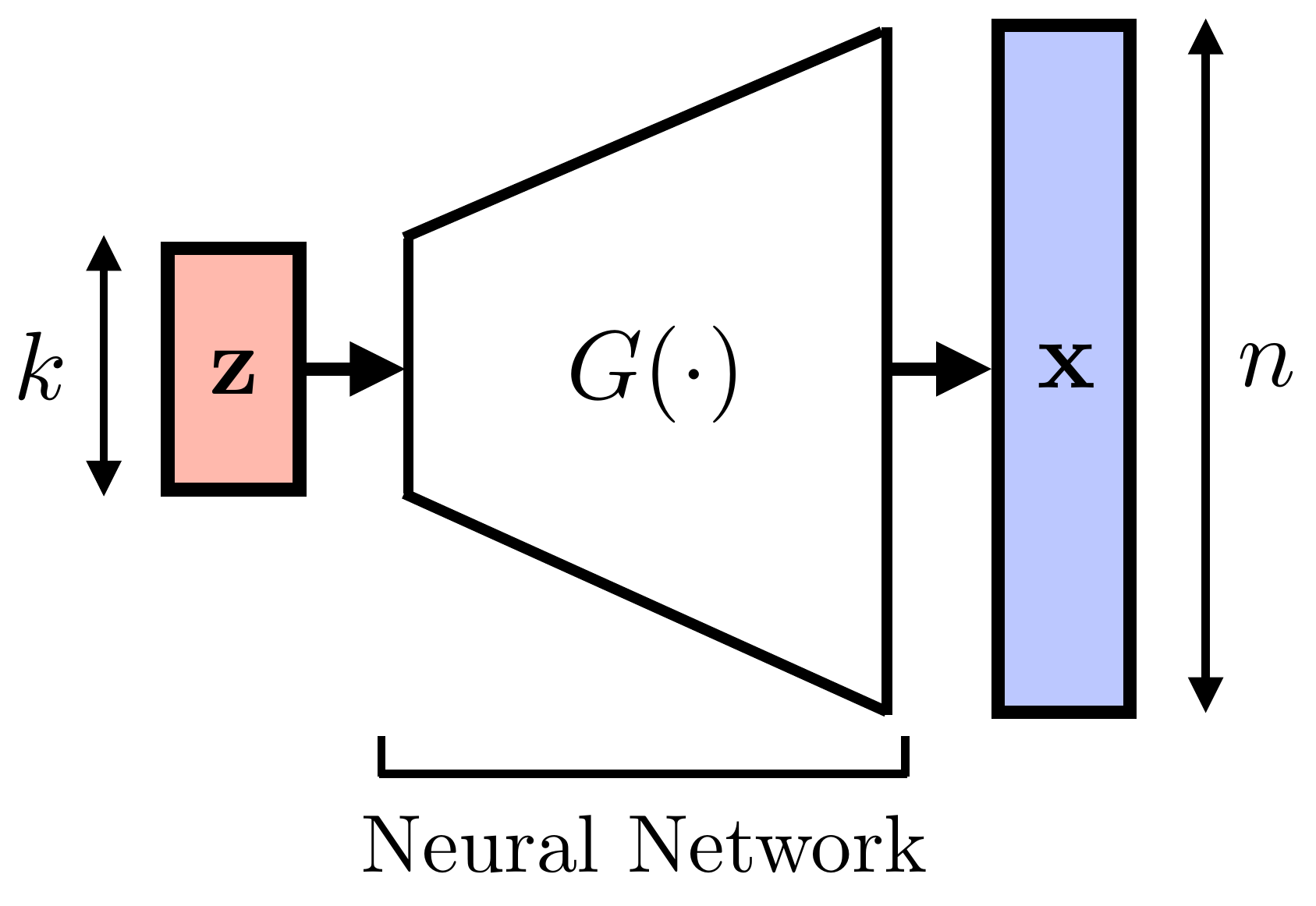}
        \par
    \end{centering}
    
    \caption{High-level structure of a typical deep generative model.  In the case of 2D images, the length-$n$ vector represents the vectorized form. \label{fig:gen_model}}
\end{figure}

The structure of a typical generative model is depicted in Figure \ref{fig:gen_model}.  The function $G$ maps a low-dimensional input $\bz \in \bbR^k$ to a high-dimensional signal $\bx \in \bbR^n$, with the internal structure of $G$ typically being a neural network.  As a toy example, with $k=1$ and $n=2$, the function
\begin{equation}
    G(z) = \big[ \sin(z), \cos(z) \big]^T
\end{equation}
maps $z \in [-\pi,\pi]$ to points on the unit circle in $\bbR^2$.  As a more realistic example, for a relatively simple data set such as MNIST, $G$ might consist of $k$ in the tens and produce $28 \times 28$ images (i.e., $n = 784$), whereas a generative model for face images might have $k$ in the hundreds, and a number of pixels in the thousands or more.  

Broadly, the use of generative priors in inverse problems consists of two main steps that are typically decoupled:
\begin{itemize}
    \item[(i)] Given suitably representative training data, train the generative model $G$ (or find a pre-trained one).
    \item[(ii)] Given the generative model $G$ and compressed measurements such as $\by=\bA\xtrue$, run an optimization procedure (e.g., \eqref{eq:x_Gz}--\eqref{eq:z_argmin} below) to produce an estimate $\hat{\bx}$ of $\xtrue$ lying in (or near) the range of $G$.
\end{itemize}
Step (i) has been widely studied in the machine learning literature, with prominent methods including generative adversarial networks \cite{goodfellow2014generative}, variational autoencoders \cite{kingma2013auto}, and so on.

At first glance, performing a theoretical analysis for signal recovery in this setup may appear to be daunting.  A typical neural network induces a highly complicated non-linear mapping; the network architecture and training algorithm may play a major role; and using training data inevitably leads to challenges relating to generalization error.

The pioneering work of Bora \emph{et al.}~\cite{bora2017compressed} circumvented these challenges by identifying simple properties of typical generative models that suffice to give meaningful recovery guarantees.  As a result, more fine-grained issues centered around training, generalization, and representation error are essentially abstracted away (though their further study would still be of significant interest).

Specifically, the following two mathematical classes of generative models were proposed in \cite{bora2017compressed}:
\begin{itemize}
    \item[(i)] $G$ is a Lipschitz continuous function, with Lipschitz constant denoted by $L$;
    \item[(ii)] $G$ is a neural network with ReLU activations,\footnote{Other piecewise linear activations can also be considered, but ReLU is of primary interest due to its widespread use in practice.} and the width and depth of the network are denoted by $w$ and $d$. 
\end{itemize}
The Lipschitz assumption can easily be shown to be satisfied by neural networks with Lipschitz activation functions (e.g., ReLU, sigmoid, and more) and bounded weights, and the ReLU network assumption is also natural in view of the ubiquity of ReLU networks in practice.  While the second class is essentially encompassed by the first, it is still of interest to study it separately, since doing so yields slightly stronger results, as well as further insights via a distinct analysis.

\subsection{Statistical Upper Bounds on the Reconstruction Error} \label{sec:ub_gen}

The following two theorems give upper bounds on the reconstruction error (in terms of the number of measurements $m$) under the Lipschitz and ReLU assumptions, respectively, considering a (possibly impractical) decoding rule based on solving a constrained $\ell_2$-minimization problem.

\begin{theorem} \label{thm:ub_lipschitz}
    {\em (Upper Bound for Lipschitz Generative Models \cite[Thm.~1.2]{bora2017compressed})}
    Let $G \colon \bbR^k \to \bbR^n$ be an $L$-Lipschitz generative model, and let the measurement matrix $\bA \in \bbR^{m \times n}$ have i.i.d.~$\calN\big(0,\frac{1}{m}\big)$ entries.  Suppose that, upon observing $\by = \bA\xtrue + \boldeta$ for some noise vector $\boldeta$, the decoder forms the estimate 
    \begin{align}
        \hat{\bx} &= G(\hat{\bz}), \quad \text{where} \label{eq:x_Gz} \\
        \hat{\bz} &= \argmin_{\zopt \in \bbR^k \,:\, \|\zopt\|_2 \le r} \|\by - \bA G(\zopt)\|_2. \label{eq:z_argmin}
    \end{align}
    Then, for any $\delta \in (0,1)$, if $m = \Omega\big( k \log \frac{Lr}{\delta} \big)$ with a sufficiently large implied constant, then it holds with probability $1 - e^{-\Omega(m)}$ that
    \begin{equation}
        \|\hat{\bx} - \xtrue\|_2 \le 6 \min_{\zopt \in \bbR^k \,:\, \|\zopt\|_2 \le r} \|G(\zopt) - \xtrue\|_2 + 3\|\boldeta\|_2 + 2\delta. \label{eq:ub_guarantee}
    \end{equation}
\end{theorem}

\begin{theorem} \label{thm:ub_relu}
    {\em (Upper Bound for ReLU Generative Models \cite[Thm.~1.1]{bora2017compressed})}
    Let $G \,:\, \bbR^k \to \bbR^n$ be a neural network with ReLU activations, width $w$, and depth $d$, and let the measurement matrix $\bA \in \bbR^{m \times n}$ have i.i.d.~$\calN\big(0,\frac{1}{m}\big)$ entries.  Suppose that, upon observing $\by = \bA\xtrue + \boldeta$ for some noise vector $\boldeta$, the decoder forms the estimate
    \begin{align}
        \hat{\bx} &= G(\hat{\bz}), \quad \text{where} \label{eq:x_Gz2} \\
        \hat{\bz} &= \argmin_{\zopt \in \bbR^k} \|\by - \bA G(\zopt)\|_2. \label{eq:z_argmin2}
    \end{align}
    Then, if $m = \Omega( k d \log w )$ with a sufficiently large implied constant, then it holds with probability $1 - e^{-\Omega(m)}$ that
    \begin{equation}
        \|\hat{\bx} - \xtrue\|_2 \le 6 \min_{\zopt \in \bbR^k} \|G(\zopt) - \xtrue\|_2 + 3\|\boldeta\|_2. \label{eq:ub_guarantee2}
    \end{equation}
\end{theorem}

We note that these results provide \emph{non-uniform} recovery guarantees, holding with respect to the randomness in $\bA$ for fixed $\xtrue$.  However, as noted in \cite[Remark 1]{liu2020information}, if there is no representation error (i.e., $\xtrue \in {\rm Range}(G)$ and the first term in \eqref{eq:ub_guarantee} or \eqref{eq:ub_guarantee2} is zero), the proofs in \cite{bora2017compressed} also provide stronger \emph{uniform} guarantees, establishing that a single matrix $\bA$ works for all $\xtrue$.

The first term in \eqref{eq:ub_guarantee} (and \eqref{eq:ub_guarantee2}) amounts to being within a constant factor of the best approximation (thus measuring the \emph{representation error}), and the second term captures the effect of noise.  The $2\delta$ term in \eqref{eq:ub_guarantee} is more subtle, and captures the fact that more measurements are needed to accurately recover details of the signal at increasingly fine scales \cite{bora2017compressed}.  In contrast, no such term is present in \eqref{eq:ub_guarantee2}.

The number of measurements above can be contrasted with the typical $O(s \log n)$ scaling for sparse priors.  The most important distinction here is not the different logarithmic terms, but rather, the fact that \emph{for accurate modeling, the required $k$ (generative priors) may be much smaller than the required $s$ (sparse priors)} due to $G$ being more targeted to the task at hand.

Slightly more general statements are given in \cite{bora2017compressed}, in which the minimization problem in \eqref{eq:z_argmin} or \eqref{eq:z_argmin2} is only solved to within $\epsilon$, and $2\epsilon$ is added to the right-hand side of \eqref{eq:ub_guarantee} or \eqref{eq:ub_guarantee2}.  While gradient-based methods can be highly effective in practice \cite{bora2017compressed}, rigorously guaranteeing $\epsilon$-optimality for small $\epsilon > 0$ may be very difficult due to the potentially complicated (e.g., highly non-convex) optimization landscape.  In Section \ref{sec:opt_gen}, we summarize some results that overcome this limitation, at the expense of imposing stronger assumptions on $G$.

\textbf{Overview of proofs:} The proofs of both Theorems \ref{thm:ub_lipschitz} and \ref{thm:ub_relu} are based on the {\em set-restricted eigenvalue condition} (S-REC), which formalizes the intuition that $\bA\bx_1$ and $\bA\bx_2$ should not be too close relative to the separation between two possible signals $\bx_1$ and $\bx_2$.  For instance, if $\bA\bx_1 = \bA\bx_2$ then clearly the two cannot be distinguished.  More generally, $\bx_1 - \bx_2$ should be \emph{far from the nullspace} of $\bA$.

\begin{definition}
    \emph{(Set-Restricted Eigenvalue Condition (S-REC) \cite[Def.~1]{bora2017compressed})}
    Fix $\calS \subseteq \bbR^n$, along with $\gamma > 0$ and $\delta \ge 0$.  The matrix $\bA \in \bbR^{m \times n}$ is said to satisfy the $\SREC(\calS,\gamma,\delta)$ if, for all $\bx_1$ and $\bx_2$ in $\calS$, it holds that $\|\bA(\bx_1-\bx_2)\|_2 \ge \gamma\|\bx_1-\bx_2\|_2 - \delta$.
\end{definition} 

Notice that this definition bounds $\|\bA(\bx_1-\bx_2)\|_2$, whereas analogous definitions based on sparsity simply bound $\|\bA\bx\|_2$ for sparse $\bx$ (e.g., see \eqref{eq:rip}).  Intuitively, this is because $\|\bA(\bx_1-\bx_2)\|_2$ is the more directly relevant quantity, but $\|\bA\bx\|$ can be used for sparse signals since the difference of two sparse signals is still sparse (unlike for general generative priors).

It is shown in \cite{bora2017compressed} that the $\SREC(\calS,\gamma,\delta)$ with $\gamma = \frac{1}{2}$, coupled with a simpler property of the form $\|\bA\bx\| \le 2\|\bx\|$ (for some fixed $\bx$), suffices to establish a recovery guarantee of the form \eqref{eq:ub_guarantee} or \eqref{eq:ub_guarantee2}, with the minimum being taken over $\calS$.  Since $\|\bA\bx\|_2 \le 2\|\bx\|_2$ holds with high probability by standard Gaussian concentration, it only remains to show that Gaussian matrices satisfy the $\SREC$ with high probability.

When $G$ satisfies the Lipschitz property (Theorem \ref{thm:ub_lipschitz}), the idea is to establish the desired behavior on a finite subset of $\calS = \{ G(\bz) \,:\, \|\bz\|_2 \le r \}$, and then transfer this to the full set.\footnote{More precisely, to avoid a worsened logarithmic factor, \cite{bora2017compressed} adopts a chaining argument that studies a \emph{sequence} of finite sets corresponding to increasingly fine scales.}  When working with a finite subset, one can study the norm-preserving properties of Gaussian matrices, as pioneered by Johnson and Lindenstrauss \cite{johnson1984extensions}.  The rough intuition behind the scaling on $m$ is that we need to cover $\calS$ such that every signal in $\calS$ is $\delta$-close to some point, and by the Lipschitz property of $G$, this amounts to similarly covering $\{\bz \in \bbR^k \,:\, \|\bz\|_2 \le r\}$ with closeness $\frac{\delta}{L}$.  This is known to be possible with a set of size $\exp\big( O\big( k\log\frac{Lr}{\delta} \big)\big)$, and the scaling on $m$ arises as the log of this size. 

For ReLU neural networks (Theorem \ref{thm:ub_relu}), the idea is that since the ReLU activation function is piecewise linear, so is the overall function $G$ (possibly with a huge number of pieces).  Within a linear region, one can again appeal to standard norm-preserving properties of Gaussian matrices, and a union bound can then be applied over all pieces.  A counting argument reveals that there are $w^{O(kd)}$ such pieces, and the bound on $m$ arises as the log of this number.

\subsection{Information-Theoretic Lower Bounds} \label{sec:lb_gen}

To assess the degree of optimality of the upper bounds, it is useful to establish \emph{information-theoretic lower bounds} (i.e., converse/impossibility results) stating that no estimation procedure can hope to improve beyond a certain limit, in terms of the estimation error and/or number of measurements.  Results of this kind were independently established by Kamath, Price, and Karmalkar \cite{kamath2020power} and Liu and Scarlett \cite{liu2020information}.

The following theorem of \cite{kamath2020power} provides such a lower bound in the case of Lipschitz continuous generative priors, and serves as a counterpart to the upper bound in Theorem \ref{thm:ub_lipschitz}.

\begin{theorem} \label{thm:lb}
    {\em (Lower Bound for Lipschitz Generative Models \cite[Thm.~1.1]{kamath2020power})}
    For any input/output sizes $k$ and $n$, and positive constants $L$, $r$, and $\delta$ such that $\log\frac{Lr}{\delta} \ge 1$, there exists a generative model $G \,:\, \bbR^k \to \bbR^n$ such that the following holds:  If there exists a random measurement matrix $\bA$ and a decoder (with access to $\bA$ and $\by = \bA\xtrue$) that is guaranteed to return $\hat{\bx}$ satisfying
    \begin{equation}
        \|\hat{\bx} - \xtrue\|_2 \le C \min_{\zopt \in \bbR^k \,:\, \|\zopt\| \le r} \|G(\zopt)- \xtrue\|_2 + \delta  \label{eq:lb_guarantee}
    \end{equation}
    with probability at least $\frac{3}{4}$ for some absolute constant $C$, then it must be the case that
    \begin{equation}
        m = \Omega\Big( \min\Big\{ k \log \frac{Lr}{\delta}, n \Big\} \Big).
    \end{equation}
\end{theorem}

This result establishes that $O\big( k \log\frac{Lr}{\delta}\big)$ is indeed the correct scaling (in the most interesting regime where this quantity is below $O(n)$), and that the additive dependence on $\delta$ in \eqref{eq:ub_guarantee} is unavoidable, unlike the case of a sparse prior (see \eqref{eq:l2l2}).  We note that this result holds for a ``worst-case'' generative model satisfying the assumptions of Theorem \ref{thm:ub_lipschitz}; it may very well be the case that further assumptions on $G$ can decrease the required $m$.

Theorem \ref{thm:lb} concerns the case that there is no noise (i.e., $\boldeta = \bzero$), but crucially relies on considering signals with representation error in order to establish the hardness result.  The opposite approach was taken in \cite{liu2020information}, in which it was assumed that there is no representation error, but that $\boldeta$ is present in the form of i.i.d.~Gaussian noise.  An analog of Theorem \ref{thm:lb} was given, though Theorem \ref{thm:lb} has the advantage of holding for general combinations of $(n,k,L,r,\delta)$, whereas \cite{liu2020information} requires $n$ to be large enough such that $\log\frac{Lr}{\delta} = O(\log\frac{n}{k})$.

An advantage of the approach in \cite{liu2020information}, on the other hand, is that it also provides a lower bound establishing conditions under which Theorem \ref{thm:ub_relu} is near-optimal, i.e., handling the specific case of ReLU generative models, and characterizing the dependence on the network depth and width.

Before stating this lower bound for ReLU networks, it is useful to highlight what the upper bound in Theorem \ref{thm:ub_relu} gives in the case of Gaussian noise and no representation error.  As stated in \cite[Cor.~2]{liu2020information}, if we have
\begin{equation}
    \xtrue \in {\rm Range}(G), \quad \text{and} \quad \boldeta \sim N\Big(\bzero,\frac{\alpha}{m} \bI_m\Big) \label{eq:simplifying}
\end{equation}
for some $\alpha > 0$, then there exists a measurement matrix\footnote{Equation \eqref{eq:mse_guarantee} also holds when $\bA$ is i.i.d.~Gaussian according to Theorem \ref{thm:ub_relu}, but it is convenient to work with fixed $\bA$ in this part.  As discussed following Theorem \ref{thm:ub_relu}, the upper bound with fixed $\bA$ crucially relies on having no representation error.  In view of this, Theorem \ref{thm:lb_relu} as stated may appear to have a weakness of only lower bounding the number of measurements under a stricter uniform recovery guarantee.  However, it is discussed in \cite[Remark 1]{liu2020information} that the proof readily provides a similar statement for non-uniform recovery.} 
$\bA \in \bbR^{m \times n}$ with squared Frobenius norm $\|\bA\|_{\rmF}^2 \le n$ such that the mean squared error is upper bounded by
\begin{equation}
    \bbE\big[ \| \hat{\bx} - \xtrue \|_2^2 \big] \le O(\alpha).  \label{eq:mse_guarantee}
\end{equation}
Intuitively, this amounts to accurately reconstructing $\bx$ with the amount of error matching the noise level.

The lower bound in this setting is more complicated than the case of Lipschitz generative models, so we provide an informal statement, and refer the reader to \cite{liu2020information} for the details.

\begin{theorem} \label{thm:lb_relu}
    {\em (Lower Bound for ReLU Networks (Informal) \cite[Thm.~7]{liu2020information})} Consider the case that $G \,:\,\bbR^k \to \bbR^n$ is a ReLU network with depth $d$ and width $w$.  Suppose that there exists a measurement matrix $\bA$ with $\|\bA\|_{\rmF}^2 \le n$ and a decoder such that when \eqref{eq:simplifying} holds, the resulting estimate $\hat{\bx}$ is guaranteed to satisfy \eqref{eq:mse_guarantee}.  Then, we have the following:
    \begin{itemize}
        \item There exists $G$ with depth $d = 2$ and large width $w$ such that it must be the case that $m = \Omega(k \log w)$.
        \item There exists $G$ with width $w = O(n)$ and large depth $d$ such that it must be the case that $m = \Omega(kd)$.
        \item There exists $G$ with simultaneously large width and large depth such that it must be the case that $m = \Omega\big(kd \frac{\log w}{\log n}\big)$.
    \end{itemize}
\end{theorem}

Observe that the number of measurements matches the $O(kd \log w)$ upper bound to within a constant factor (first case) or an $O(\log n)$ factor (second and third cases).  We note that certain cases are known where the logarithmic factor in the upper bound can be slightly reduced \cite{naderi2021beyond}.


\textbf{Overview of proofs:} As is common in proving information-theoretic lower bounds, the high-level idea behind Theorems \ref{thm:lb} and \ref{thm:lb_relu} is to establish that the relevant recovery guarantee implies being able to reliably distinguish certain well-separated signals.  If there are many such signals, then reliably distinguishing them amounts to learning a certain amount of information, and since each measurement only provides a limited amount of information, a lower bound on the number of measurements follows.

In \cite{kamath2020power}, the details are based on a reduction to communication complexity.  A subset $\calX_0$ of well-separated binary-valued signals is formed with $\log|\calX_0| = \Omega\big(\min\big\{ k \log \frac{Lr}{\delta}, n \big\}\big)$, and $\bx$ is restricted to be a weighted linear combination of several such signals plus a small Gaussian perturbation.  A communication game is set up in which one party wishes to identify one of the binary-valued signals, and for which a lower bound on the number of bits transmitted is known for achieving constant-probability success.  It is shown that transmitting a fine discretization of $\by = \bA \xtrue \in \bbR^m$ suffices for such success, from which a lower bound on $m$ follows.

\begin{figure}
    \begin{centering}
        \includegraphics[width=0.95\columnwidth]{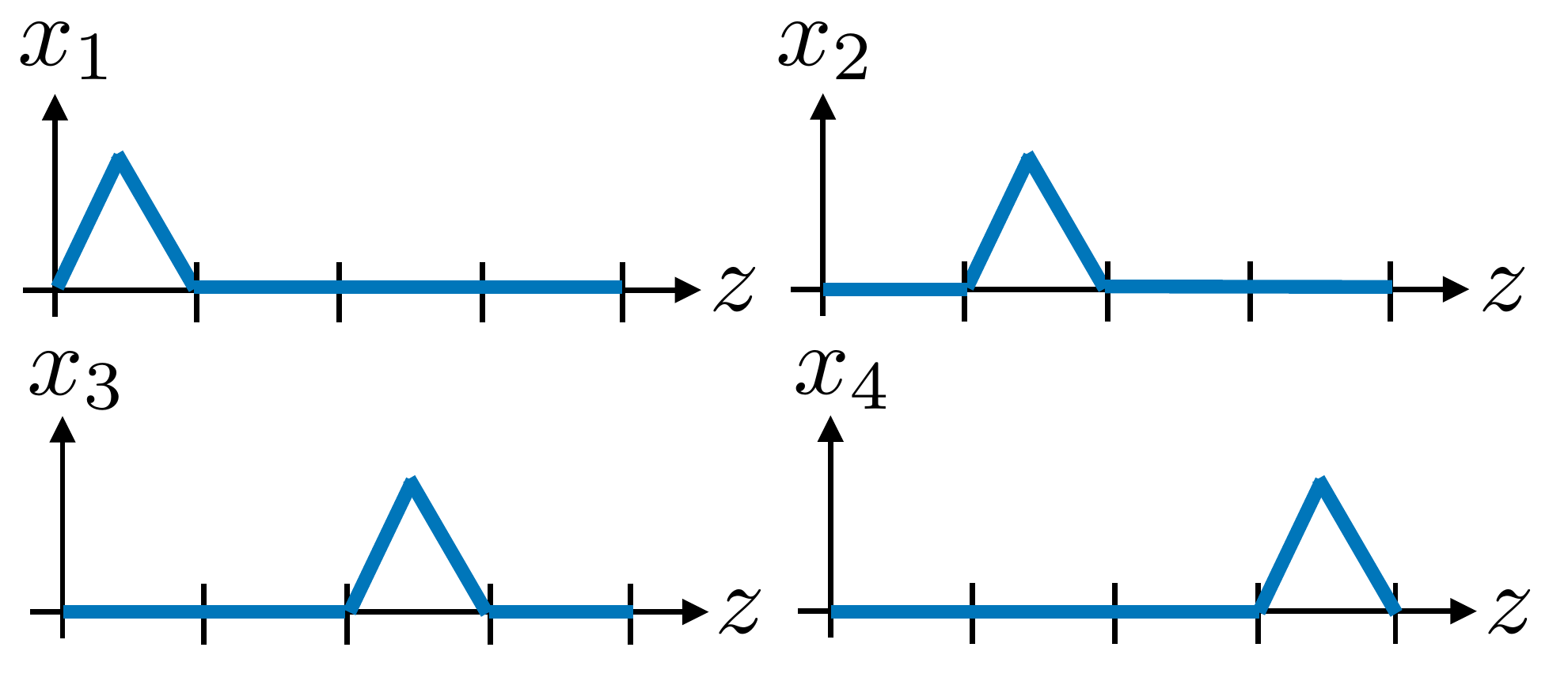}
        \par
    \end{centering}
    
    \caption{Example function mapping $z \in \bbR$ to $\bx = (x_1,x_2,x_3,x_4) \in \bbR^4$ such that the resulting signal is 1-sparse (or is the zero vector). \label{fig:four_sparse}}
\end{figure}

In \cite{liu2020information}, to prove Theorem \ref{thm:lb_relu} and a counterpart to Theorem \ref{thm:lb}, a different approach is taken.  The idea is to construct a generative model $G$ that produces \emph{sparse signals},\footnote{The ability of ReLU networks to produce sparse signals was also noted in \cite{kamath2020power}, but no analog of Theorem \ref{thm:lb_relu} was sought.} and then apply standard lower bounding techniques (e.g., based on Fano's inequality) that characterize the hardness of sparse recovery.  By studying the Lipschitz constant and/or the depth and width of $G$, and combining these with the relevant lower bounds for sparse recovery, the desired results follow.  An illustration of why neural networks can produce sparse signals is shown in Figure \ref{fig:four_sparse}; the piecewise linear functions can readily be implemented using ReLU networks.  An analog of Theorem \ref{thm:ub_lipschitz} is obtained by forming a network that produces $k$-sparse signals (with input $\bz \in \bbR^k$), whereas Theorem \ref{thm:ub_relu} is based on producing $kk_0$-sparse signals with $k_0 > 1$, using recursively-defined mappings that operate at $k_0$ different scales. 

We refer the reader to \cite{kamath2020power,liu2020information} for the full details of the above proof outlines.

\subsection{Optimization Guarantees for Random Generative Priors} \label{sec:opt_gen}

As we mentioned above, finding an optimal or near-optimal solution to problems such as \eqref{eq:z_argmin} and \eqref{eq:z_argmin2} may not be possible with an efficient algorithm.  Thus, there is substantial motivation to give recovery guarantees for specific tractable optimization procedures (which comes at the expense of stronger assumptions on $G$).  In this subsection, we outline some examples of such guarantees.

We again consider $G \colon \bbR^k \to \bbR^n$ being a ReLU neural network, but now with two main additional assumptions, namely, (i) sufficient expansivity (i.e., increase in the number of nodes) from layer to layer, and (ii) random Gaussian network weights.  Due to the second assumption, such networks would not produce meaningful signals in practice. 
However, as noted in \cite{hand2018global}, some trained networks do exhibit Gaussian-like statistics, and more importantly, understanding random networks is already highly challenging and serves as a good starting point towards increasingly more realistic scenarios.

We focus primarily on the results of Hand and Voroninski \cite{hand2018global} and Huang \emph{et al.}~\cite{huang2021provably}.  It was first shown in \cite{hand2018global} that that the optimization landscape in formulation \eqref{eq:z_argmin2} is favorable for gradient algorithms if the network architecture satisfies certain deterministic properties and if there are a sufficient number of random Gaussian measurements.  Inspired by this landscape, \cite{huang2021provably} introduced a specific subgradient algorithm that provably converges.  It was additionally established in \cite{hand2018global} that under the above-mentioned assumptions of expansivity and random weights, the desired deterministic properties are satisfied with high probability.  The assumptions made were then relaxed in various subsequent works \cite{daskalakis2020constant, joshi2021plugin, cocola2022signal}, several of which we will discuss in Section \ref{sec:stat_further}.

\subsubsection{Model for G}
We consider a generator $G: \bbR^k \to \bbR^n$ given by a $d$-layer fully connected neural network with ReLU activations and no bias terms.  That is, 
\begin{align}
G(\bz) = \relu(\bW_d \ldots \relu(\bW_2 \relu( \bW_1 \bz)) \ldots ), \label{defn-g}
\end{align}
where $\relu(\cdot) = \max\{\cdot, 0\}$ applies entry-wise, $\bW_i \in \R^{n_i \times n_{i - 1}}$ for $i = 1,\dotsc,d$,  and $n_0 = k$  and $n_d=n$.  

\subsubsection{Deterministic conditions used in the analysis}

Here we present two useful deterministic conditions on the generative model and measurement model.  The results to follow will show that these deterministic conditions are sufficient for certain recovery guarantees, and are satisfied with high probability for i.i.d.~Gaussian distributions on $G$ and $\bA$.

The first condition is the Weight Distribution Condition (WDC), which applies to individual weight matrices $\bW_i$.  

\begin{definition}[Weight Distribution Condition (WDC) \cite{hand2018global}] \label{GA:df1}
    A matrix $\bW \in \mathbb{R}^{\kappa \times \ell}$ satisfies the \emph{Weight Distribution Condition} with constant $\epsilon$ if, for all non-zero $\bu, \bv \in \mathbb{R}^\ell$, it holds that
    \begin{align}
    \left\| \sum_{i = 1}^{\kappa} \bone_{\bw_i \cdot \bu > 0} \bone_{\bw_i \cdot \bv > 0} \cdot \bw_i \bw_i^T - \bQ_{\bu, \bv} \right\|_2 \leq \epsilon, & \nonumber \\ \hbox{ with } \bQ_{\bu, \bv} = \frac{\pi - \theta}{2 \pi} \bI + \frac{\sin \theta}{2 \pi} \bM_{\bu, \bv},&
    \end{align}
    where $\bw_i^T \in \mathbb{R}^\ell$ is the $i$-th row of $\bW$; $\theta$ is the angle between $\bu$ and $\bv$; $\bM_{\bu, \bv} \in \mathbb{R}^{\ell \times \ell}$ is the matrix that maps $\frac{\bu}{\|\bu\|_2} \mapsto \frac{\bv}{\|\bv\|_2}$, $\frac{\bv}{\|\bv\|_2} \mapsto \frac{\bu}{\|\bu\|_2}$, and $\bt \mapsto 0$ for all $\bt$ orthogonal to $\sspan(\{\bu, \bv\})$; and $\bone_S$ is the indicator function on $S$.
\end{definition}
This condition can be viewed as a generalization of an approximate isotropy condition; for example, if $\bu = \bv$, the condition states that $\sum_{i = 1}^{\kappa} \bone_{\bw_i \cdot \bu > 0} \bone_{\bw_i \cdot \bv > 0} \cdot \bw_i \bw_i^T$ is close to $\frac{1}{2}\bI$.  The indicator functions in the summation arise from taking the derivative of the ReLU function.  

The second condition is the Range Restricted Isometry Condition (RRIC), which applies to the pair $(G, \bA)$.

\begin{definition}[Range Restricted Isometry Condition (RRIC) \cite{hand2018global}] \label{GA:df2}
A matrix $\bA \in \mathbb{R}^{m \times n}$ satisfies the \emph{Range Restricted Isometry Condition} with respect to $G$ with constant $\epsilon$ if, for all $\bz_1, \bz_2, \bz_3, \bz_4 \in \mathbb{R}^k$, it holds that
\begin{align}
\Big| \langle \bA(&G(\bz_1) - G(\bz_2)), \bA(G(\bz_3) - G(\bz_4)) \rangle \nonumber \\ &- \langle G(\bz_1) - G(\bz_2), G(\bz_3) - G(\bz_4) \rangle \Big| \nonumber  \\
&\leq \epsilon \|G(\bz_1) - G(\bz_2)\|_2 \|G(\bz_3) - G(\bz_4)\|_2.
\end{align}
\end{definition}
This condition states that $\bA$ acts like an isometry when acting on pairs of secant directions (i.e., differences of two signals) with respect to the range of $G$.

\subsubsection{Favorable landscape for compressive sensing with gradient algorithm under deterministic conditions}

Under the deterministic conditions given above, it can be established that the loss landscape is favorable for optimization.  Consider a signal given by $\xtrue = G(\zo)$ for some $\zo$, and let the measurement vector be $\by = \bA \xtrue + \boldeta$ with i.i.d.~Gaussian $\boldeta$.  We are interested in the optimization problem
\begin{equation}
\min_{\zopt} f(\zopt), \quad f(\zopt) := \| \bA G(\zopt) - \by\|_2^2.
\end{equation}
The following result shows that under the WDC and RRIC, $f$ does not have any spurious local minima outside of $\bz$ and a negative multiple of $\bz$.  Here and subsequently, when we write ${\rm poly}(d)$, we mean that the result holds true when this is replaced by $d^c$ for a suitable constant $c > 0$ (possibly differing in each occurrence).  In addition, we let $D_{\bv}f(\bz)$ denote the directional derivative with direction $\bv \in \bbR^k$, and let $\calB(\bz,r)$ denote the radius-$r$ ball centered at $\bz$.

\begin{theorem} \label{thm-multi-layer-deterministic}
\emph{(Favorable Optimization Landscape \cite[Thm.~4]{hand2019globalieee}) }
Fix $\eps >0$ such that $K_1 {\rm poly}(d) \eps^{1/4} \leq 1$, and let $d \geq 2$.  Suppose that $G$ is such that $\bW_i$ satisfies the  WDC with constant $\eps$ for all $i = 1, \dotsc, d$, and that $\bA$ satisfies the RRIC with respect to $G$ with constant $\eps$.  Then, for all non-zero $\zopt$ and $\zo$, there exists $\vzoptzo \in \R^k$ such that the one-sided directional derivatives of $f$ satisfy
\begin{align}
&D_{-\vzoptzo} f(\zopt) < -K_3 \frac{\sqrt{\eps}\, {\rm poly}(d)}{2^d} \max\big\{ \|\zopt\|_2, \|\zo\|_2 \big\}, \quad \\
&D_\bt f(\bzero) < - \frac{1}{8 \pi 2^d} \|\zo\|_2, \nonumber  \\
&\forall \bt \neq 0, \zopt  \not\in  \{ \bzero \} \cup \calB(\zo, K_2 {\rm poly}(d) \eps^{1/4} \|\zo\|_2) \nonumber  \\ &\qquad \qquad \qquad \ \cup \calB(-\rho \zo, K_2 {\rm poly}(d) \eps^{1/4} \|\zo\|_2),
\end{align}
where $\rho = \rho_d$ is a positive number that converges to $1$ as $d \to \infty$,  and $K_1$, $K_2$, and $K_3$ are universal constants.
\end{theorem}

While the above expressions are somewhat technical, the simple idea is that except for points close to $\bz^*$ and $-\rho \bz^*$, we have a negative upper bound on the directional derivative, which precludes spurious minima.  Moreover, the radius around $\bz^*$ and $-\rho \bz^*$ becomes arbitrarily small as $\epsilon$ decreases.

There is an explicit formula for $\vzoptzo$, given by
\begin{align}
\vzoptzo &= \begin{cases}\nabla f(\zopt) & \text{differentiable at $\zopt$,}\\ \mathop{\lim_{\delta \downarrow 0}} \nabla f(\zopt + \delta \bz') & \text{otherwise,} \end{cases} 
\end{align}
where $\bz'$ can be arbitrarily chosen such that $G$ is differentiable at $\zopt+\delta \bz'$ for sufficiently small $\delta$.  Such a $\bz'$ exists by the piecewise linearity of $G$, and can be generated randomly with probability one.

Note that the dependence on $2^d$ in the bounds is an artifact of the underlying scaling of $f(\zopt)$, and does not indicate a vanishingly small derivative.  Roughly speaking, the ReLU activation functions zero out around half of its arguments.  Hence, while $\bW_i$ has spectral norm approximately one, the rows of $\bW_i$ that are retained by the ReLU will have spectral norm approximately $\frac{1}{2}$.  Thus, $f(\zopt)$ itself is on the order of $2^{-d}$ under the RRIC and WDC for appropriately small $\eps$.

\subsubsection{Subgradient algorithm and convergence guarantee under deterministic conditions}
Building on Theorem \ref{thm-multi-layer-deterministic}, \cite{huang2021provably} proposed a subgradient algorithm and showed that it has a rigorous convergence guarantee.  Since the cost function $f(\zopt)$ is continuous, piecewise quadratic, and not differentiable everywhere, the algorithm is defined with respect to a generalized gradient, called the Clarke subdifferential, generalized subdifferential, or generalized subgradient (e.g., see \cite{clason2017nonsmooth} for the definition).

The algorithm operates as follows given some initialization:
\begin{itemize}
    \item Compute a vector in the subgradient of the objective at the current iterate;
    \item Update the current position using the subgradient and a fixed step size;
    \item If negating the current iterate reduces the value of the objective, then do so;
    \item Repeat until a stopping criterion is met.
\end{itemize}
Note that the third step is non-standard, and is motivated by the landscape properties stated in Theorem \ref{thm-multi-layer-deterministic}.

\begin{theorem} \label{GA:th1}
\emph{(Optimization Guarantee \cite[Thm.~1]{huang2021provably}) }
Suppose that the WDC and RRIC
hold with $\epsilon \leq \frac{C_1}{{\rm poly}(d)}$, and the noise $\boldeta$ satisfies $\|\boldeta\|_2 \leq \frac{C_2 \|\zo\|_2}{{\rm poly}(d) 2^{d/2}}$. Consider the iterates $\{\zopt_t\}$ generated by the preceding algorithm with step size $\nu = C_3 \frac{2^d}{{\rm poly}(d)}$.  There exists a number of iterations, denoted by $\tau$ and  upper bounded by $\tau \leq \frac{C_4 f(\bz_0) 2^d}{{\rm poly}(d) \epsilon \|\zo\|_2}$ when the initialization is $\bz_0$, such that
\begin{equation}\label{GA:e63}
\|\zopt_{\tau} - \zo\|_2 \leq C_5 {\rm poly}(d)\sqrt{\epsilon} \|\zo\|_2  + C_6 {\rm poly}(d) 2^{d/2} \|\boldeta\|_2.
\end{equation}
In addition, for all $t \geq \tau$, we have
\begin{align}
\|\zopt_{t + 1} - \zo\|_2 &\leq C^{t + 1 - \tau} \|\zopt_{\tau} - \zo\|_2 + C_7 2^{d/2} \|\boldeta\|_2, \label{GA:e68}
\end{align}
and
\begin{align}
\|G(\zopt_{t + 1}) - G(\zo)\|_2 &\leq \frac{1.2}{2^{d/2}} C^{t + 1 - \tau} \|\zopt_{\tau} - \zo\|_2 + 1.2 C_7 \|\boldeta\|_2, \label{GA:e74}
\end{align}
where $C = 1 - \frac{\nu}{2^d} \frac{7}{8} \in (0, 1)$.
Here, $C_1, \ldots, C_7$ are universal positive constants.
\end{theorem}

In accordance with the above discussion on the dependence on $2^d$, the initial value $f(\bz_0)$ scales with respect to $d$ as $2^{-d}$ under the WDC and RRIC.  Hence, and in view of the assumption $\epsilon \leq \frac{C_1}{{\rm poly}(d)}$, we find that Theorem \ref{GA:th1} establishes that after a number of iterations that is polynomial in $d$, the modified subgradient algorithm converges linearly to $\zo$, up to the noise level.  Note that this convergence guarantee applies for an arbitrary initialization, though more iterations may be required for initializations that are large in norm.  As with Theorem \ref{thm-multi-layer-deterministic}, while the landscape is non-convex, the theorem establishes under the WDC and RRIC that the non-convexity is mild and does not result in spurious minima, except $ -\rho \bz^*$ which the above algorithm avoids. 

\subsubsection{Random $G$ satisfies the deterministic conditions with high probability}

Finally, the following result establishes that the WDC (Definition \ref{GA:df1}) and RRIC (Definition \ref{GA:df2}) are satisfied with high probability provided that (i) $G$ has i.i.d.~Gaussian weights and is sufficiently expansive, and (ii) $\bA$ has i.i.d.~Gaussian entries and sufficiently many rows.

\begin{proposition} \label{thm-multi-layer}
\emph{(High-Probability Behavior of Random Models \cite[Prop.~6]{hand2019globalieee}) }
Fix $0 < \eps < 1$.  Assume that $G$ follows the structure in \eqref{defn-g} with $n_i \geq c n_{i-1} \log n_{i-1}$ for all $i = 1, \dotsc, d$, and that $m > c  k d \log \Pi_{i=1}^d n_i$.  Moreover, assume that the entries of $\bW_i$ are i.i.d.~$\calN\big(0, \frac{1}{n_i}\big)$, and the entries of $\bA$ are i.i.d.~$\calN\big(0, \frac{1}{m}\big)$.  
Then, $\bW_i$ satisfies the WDC with constant $\eps$ for all $i$ and $\bA$ satisfies the RRIC with respect to $G$  with constant $\eps$ with probability at least $1 - O\big( \sum_{i=1}^d  n_i e^{-\gamma n_{i-1} }-  e^{-\gamma m}\big)$.  Here $c$ and $\gamma^{-1}$ are constants that depend polynomially on $\eps^{-1}$.
\end{proposition}

Observe that the leading term in the number of measurements is $kd$, as is the case in Theorem \ref{thm:ub_relu}.  However, depending on the expansivity, the logarithmic term $\log \Pi_{i=1}^d n_i$ can be order-wise larger than $\log w$, and accordingly could potentially be improved.

The proof of Proposition \ref{thm-multi-layer} relies on tools from non-asymptotic random matrix theory \cite{Vershynin2012}.  Typically, establishing a matrix concentration result like the WDC with high probability would involve three steps: showing high probability concentration of the matrix applied to fixed vectors $(\bu, \bv)$, bounding an appropriate Lipschitz constant, and taking a union bound over a net whose size depends on that Lipshitz constant.  Because the matrix in the WDC is discontinuous with respect to $(\bu, \bv)$, this approach must be modified.  The authors of \cite{hand2019globalieee} show that the discontinuity can be smoothed to provide a semidefinite upper bound on the desired expression, and also smoothed to provide a semidefinite lower bound.  Each of these can then be controlled by the standard approach mentioned above. 

Along similar lines to the proof of Theorem \ref{thm:ub_relu}, establishing the RRIC involves showing that the output of linear maps with ReLU activations live in a union of linear spaces, and counting the number of such subspaces.

\subsection{Further Developments} \label{sec:stat_further}

In this subsection, we provide several examples of follow-up theoretical results related to those outlined above.  We keep this summary brief, and refer the reader to the references given for further details. 

\subsubsection{Statistical Guarantees} Some further developments related to the results in Section \ref{sec:ub_gen} (and to a lesser extent, Section \ref{sec:lb_gen}) are outlined as follows.

\textbf{Mitigating representation error.} While generative priors have clear benefits over conventional priors, they can suffer from the issue of \emph{representation error}: If the signal $\xtrue$ is not exactly in the range of $G$, then an optimization procedure such as \eqref{eq:z_argmin} will always incur some amount of error no matter how many measurements $m$ we take.  In contrast, it is straightforward to devise sparsity-based solutions that are guaranteed to become arbitrarily accurate as $m$ increases.  To overcome this limitation, \cite{dhar2018modeling} proposed to model $\xtrue$ as the sum of a generative component and a sparse component, and gave a theoretical guarantee that combines the features of Theorem \ref{thm:ub_lipschitz} and analogous sparse recovery results.

With a similar motivation but a very different approach, various methods were proposed in \cite{hussein2020image,daras2021intermediate,gunn2022regularized,daras2022score} based on optimizing intermediate layers in the neural network defining $G$, which helps to expand the range of the generator and mitigate representation error.  Conditions were given under which the required number of measurements is provably smaller than Theorem \ref{thm:ub_relu}, and improvements in the out-of-distribution robustness were observed experimentally. 

\textbf{Non-linear measurement models.} While Theorem \ref{thm:ub_lipschitz} concerns linear observation models, analogous guarantees have been provided for a variety of non-linear measurement models, including 1-bit observations \cite{liu2020sample,qiu2020robust,jiao2021just}, spiked matrix models \cite{aubin2019spiked,cocola2022signal}, phase retrieval \cite{liu2021towards,hand2018phase}, principal component analysis \cite{liu2022generative}, and general single-index models \cite{liu2020generalized,liu2022noniterative,liu2022projected}.  While these each come with their own challenges, the intuition behind their associated results is often similar to that discussed above for the linear model, with the $m = O\big(k \log \frac{Lr}{\delta}\big)$ scaling typically remaining. 

\textbf{Robustness to outliers.} Theorem \ref{thm:ub_lipschitz} is primarily suited to well-behaved noise, such as Gaussian or sub-Gaussian.  In contrast, heavy-tailed noise with large outliers can considerably worsen the performance, both in theory (e.g., due to the size of $\|\boldeta\|_2$ in \eqref{eq:ub_guarantee}) and in practice (e.g., since \eqref{eq:z_argmin} is not a robust objective).  Algorithms and theory addressing this challenge were given in \cite{wei2019statistical,jalal2020robust}.  Briefly, using robust estimation techniques, one can attain an analog of Theorem \ref{thm:ub_lipschitz} even when a constant fraction of the data is drawn from a heavy-tailed distribution that yields large outliers.  See also \cite{yi2018outlier} for a theoretical study of outlier detection using generative models.

\textbf{General probabilistic priors.} Theorems \ref{thm:ub_lipschitz} and \ref{thm:ub_relu} treat $G$ as a fixed function satisfying certain properties, without addressing the fact that even over the range of $G$, some signals may be more likely than others.  This distinction is particularly important when it comes to generative models that fail to satisfy $k \ll n$ (e.g., invertible generative models with $k=n$ \cite{asim2020invertible}).  To address this, compressive sensing with general probabilistic priors was studied in \cite{jalal2021instance}.  Analogous to how covering properties play a key role in the proof of Theorem \ref{thm:ub_lipschitz}, it was shown that a \emph{probabilistic} form of covering dictates the required number of measurements.  Moreover, it was shown that in broad scenarios, using i.i.d.~Gaussian measurements and letting $\hat{\bx}$ be a random sample from the posterior of $\xtrue$ is near-optimal for estimation. 

\subsubsection{Optimization Guarantees} Some further developments related to the results in Section \ref{sec:opt_gen} are outlined as follows. 

\textbf{Weakening the expansivity condition.} 
In Proposition 1, the WDC and RRIC were established with high probability in the case of layer-wise expansivity, that is, $n_i \geq c n_{i-1} \log n_{i-1}$.  This assumption was weakened in \cite{daskalakis2020constant} to $n_i \geq c n_{i-1}$ by introducing the notion of pseudo-Lipschitzness and by placing nets over spheres in a suitably non-uniform manner.

Subsequently, it was shown in \cite{joshi2021plugin} and \cite{cocola2022signal} that layer-wise expansivity is not necessary.  Specifically, the recovery guarantee is possible even if some layers are contractive (i.e., they have fewer outputs than inputs), provided that all layers are sufficiently large relative to the input dimensionality $k$.   This is shown in \cite{joshi2021plugin} in the case of a modified gradient algorithm, and \cite{cocola2022signal} observed that the WDC of a given layer only needs to hold restricted to the range of previous layers.

\textbf{Alternative architectures.} 
The model \eqref{defn-g} assumes that the architecture of the neural network $G$ is fully-connected.  In \cite{ma2018invertibility}, a similar recovery guarantee was established in the case that $G$ has a convolutional architecture. 

\textbf{Other inverse problems.} 
Signal recovery guarantees with random generative priors have been established for a variety of inverse problems, including denoising, blind demodulation, phase retrieval, and spiked matrix models.

Various results on denoising can be found in \cite{hand2018global,lei2019inverting,aberdam2020and,heckel2021rate}.\footnote{These works also study the question of when $\bz^*$ can be recovered from $G(\bz^*)$ even in the absence of noise (e.g., see \cite{lei2019inverting} for an NP-hardness result).}  In particular, it is shown in \cite{heckel2021rate} that solving the optimization problem $\min_{\zopt} \|\by - G(\zopt)\|_2^2$ with a gradient method yields an optimal denoising rate of $O\big(\frac{k}{n}\big)$ provided that the noise is sufficiently small and the generative model has Gaussian weights and is sufficiently expansive.  An alternative approach that avoids the need for random weights and expansivity is given in \cite{aberdam2020and}, instead considering sparsity properties of the hidden layers (resulting from $\relu(z) = 0$ for $z \le 0$).

In the case of phase retrieval with random weights and expansivity, solving the optimization problem $\min_{\zopt} \| \by - | \bA G(\zopt) | \|_2^2$ allows for signal recovery with $m$ being proportional to $k$ (ignoring the $n$ and $d$ dependence) \cite{hand2018phase}.  
This dependence is information-theoretically optimal, and it is noteworthy that it is attained with an efficient algorithm under random generative priors.  In contrast, for sparse priors, there is no known practical algorithm that achieves a recovery guarantee with a linear dependence on the sparsity $s$, even though doing so is known to be information-theoretically possible.  See also \cite{hand2021optimalsample} for simplified arguments in the case of phase retrieval without prior information (i.e., general signals in $\bbR^n$).

Random generative priors have also allowed for recovery results for the case of spiked matrix models \cite{aubin2019spiked,cocola2022signal}.  The number of measurements is again shown to be information-theoretically order-optimal using an efficient algorithm, unlike in the case of sparse models.

\textbf{Other optimization algorithms.} Analogous guarantees to Theorem \ref{thm:ub_relu} were given in \cite{shah2018solving,peng2020solving} for projected gradient descent, which alternates between gradient steps and projections onto the range of $G$.  However, a notable limitation of such guarantees is that the projection step itself depends on the landscape of $G(\bz)$, and may accordingly be intractable.  The results in Section \ref{sec:opt_gen} overcome this limitation at the expense assuming random weights (along with expansivity).  Two further works gave guarantees that require neither random weights nor exact projections, but instead adopt further deterministic assumptions on $G$, roughly amounting to certain forms of smoothness.  Specifically, \cite{gomez2019fast} studied an algorithm based on Alternating Direction Method-of-Multipliers (ADMM), and \cite{nguyen2021provable} studied an algorithm based on Langevin dynamics.  Algorithms based on Langevin dynamics have also been explored in several other works on inverse problems and beyond, e.g., see \cite{jalal2021instance} for its use in posterior sampling with general probabilistic priors, and \cite{raginsky2017non} for a general non-asymptotic analysis under non-convex objectives.

Another important class of algorithms uses \emph{approximate message passing} (AMP), which is a powerful technique that has been utilized extensively in high-dimensional statistics \cite{feng2021unifying}.  Variants of AMP have successfully been devised with theoretical guarantees in several inverse problems with generative priors, including linear forward models \cite{metzler2017learned,fletcher2018inference}, spiked matrix recovery \cite{aubin2019spiked}, and phase retrieval \cite{aubin2019precise}.  Similar to Section \ref{sec:opt_gen}, these results consider generative models with random weights and architectural assumptions such as expansivity.  A key advantage of AMP is that its analysis is often powerful enough to attain precise constant factors, unlike typical analyses of gradient descent algorithms. 

\subsection{Ongoing Challenges}

Compared to explicit priors such as sparsity and low rankness, the study of generative priors remains in its relatively early stages.  In this subsection, we overview some of the ongoing challenges and open problems that may be considered in future work.

\textbf{Generative model properties.} While the Lipschitz constant and the depth/width are natural parameters to consider for the generative model, these are ``global'' properties that may not fully capture the precise structure imposed by typical generative priors.  For instance, even if the global Lipschitz constant is huge, it may be that the function is mostly sufficiently smooth to ensure that few measurements suffice.  In view of this, it would be of significant interest to identify additional properties that more precisely dictate the required number of measurements.

\textbf{Structured measurements.} Studies of compressive sensing with generative priors have predominantly focused on i.i.d.~Gaussian measurement matrices.  Non-Gaussian i.i.d.~designs have also been considered \cite{jalal2020robust}, as well as certain classes of dependent measurements \cite{naderi2021beyond}.  However, theory is still largely lacking for several kinds of measurements that are used in practice; for instance, in applications such as medical imaging, one is confined to using subsampled Fourier measurements due to the inherent design of the hardware.  Very recently, a study of such settings with generative priors was initiated \cite{berk2023coherence}, and it was demonstrated that the required number of measurements can be characterized by a coherence parameter measuring the interplay between the range of the generative model and the measurement matrix.

\textbf{Optimization guarantees with milder assumptions.}  Regarding the optimization results outlined in Section \ref{sec:opt_gen} and the related follow-up works, perhaps the most significant ongoing challenge is to expand the applicability of the theory beyond the case of random generative models, and more generally, to give analogous guarantees with as few restrictive assumptions on $G$, $\bA$, and $\boldeta$ as possible.

\textbf{Constant factors.} As exemplified in the results that we stated, most existing works on the theory of inverse problems with generative models have typically sought to characterize the scaling laws of the number of measurements, and not the finer question of precise constants.  As discussed above, progress has been made in addressing this question using approximate message passing (AMP), but broadly speaking, there remains substantial room for progress in understanding the constant factors associated with bounds on the number of measurements with generative priors.

\textbf{Role of training data.} As we discussed earlier, the consideration of properties such as the Lipschitz constant essentially abstracts away the complicated details of how the generative model was trained.  On the other hand, to attain a more complete picture of the entire learning and information processing pipeline, a refined theory might explicitly incorporate such aspects, e.g., explicitly quantifying notions such as representation and generalization, and unifying such considerations with the number of measurements in the inverse problem, the optimization algorithm used for decoding, and so on.  While a completely holistic theory may be challenging, future research could potentially take gradual steps towards this.

\textbf{Out-of-distribution performance.} One of the main potential concerns of generative priors is that they may perform poorly under distribution shift, i.e., when the training data is not fully representative of the actual signal being recovered.  Various works have started to address this limitation (see Section \ref{sec:stat_further}), but overall, we believe that it remains under-explored relative to its importance.
    
\newcommand{\reals}{\mathbb R}
\newcommand{\mB}{\mathbf{B}} 
\newcommand{\mC}{\mathbf{C}}
\newcommand{\mU}{\mathbf{U}}
\newcommand{\mA}{\mathbf{A}}
\newcommand{\mI}{\mathbf{I}}
\newcommand{\vy}{\mathbf{y}}
\newcommand{\vw}{\mathbf{w}}
\newcommand{\vv}{\mathbf{v}}

\newcommand{\vx}{\mathbf{x}}
\newcommand{\vz}{\mathbf{z}}

\newcommand{\veta}{{\bm \eta}}
\newcommand{\vtheta}{{\bm \theta}}
\newcommand{\mW}{{\bm W}}

\newcommand{\cn}{\mathrm{cn}}
\newcommand{\kout}{q}
\definecolor{DarkBlue}{rgb}{0.1,0.1,0.5}
\definecolor{BrickRed}{RGB}{203,65,84}

\section{Untrained Neural Network Priors} \label{sec:untrained} 

In this section, we consider untrained neural network priors, which, in contrast to the pre-trained generative priors considered above, work without any training data and solely based on the network architecture and the choice of optimization procedure for fitting the signal/image at inference time.  For instance, one of the earliest such techniques called Deep Image Prior (DIP) \cite{ulyanov_DeepImagePrior_2018} works by fitting a standard convolutional auto-encoder (the popular U-net~\cite{ronneberger_UNetConvolutionalNetworks_2015}) to a single noisy image via gradient descent, and regularizing by early stopping.
Untrained networks have emerged as a highly successful alternative to data-driven methods, yielding excellent performance for a variety of problems, including denoising~\cite{ulyanov_DeepImagePrior_2018,heckel_DeepDecoderConcise_2019} and compressive sensing~\cite{yoo_TimeDependentDeepImage_2021,vanveen_CompressedSensingDeep_2018,jagatap_AlgorithmicGuaranteesInverse_2019,bostan_DeepPhaseDecoder_2020,wang_PhaseImagingUntrained_2020,hyder_GenerativeModelsLowDimensional_2020}. 

Despite being neural network based, this class of methods is conceptually related to sparsity-based methods, in that it is not data-driven and it relies on broader properties of signals/images (e.g., smoothness) rather than capturing the behavior of highly specific data distributions.  Note that data is typically still used for hyperparameter tuning, similarly to sparsity-based methods.

On the other hand, untrained networks can provide significant improvements over sparsity-based methods; for example, they can give better image quality for accelerated magnetic resonance imaging~\cite{darestani_AcceleratedMRIUnTrained_2021}.  While the precise reason for this is difficult to pinpoint, it may result from the architectures of untrained networks (e.g., incorporating operations such as convolution) being able to represent typical signals/images more effectively than sparsity-based priors (e.g., dictated by low total-variation norm).


In this section, we first discuss how signal recovery can be performed using untrained neural networks, and then overview the existing theory behind this approach.

\subsection{Background}

Consider the problem of reconstructing a signal $\xtrue \in \reals^n$ from noisy linear measurements, $\vy = \mA \xtrue + \veta \in \reals^m$. The signal is often an image, in which case these equations correspond to its vectorized form. 

We let $G \,:\, \reals^p \to \reals^n$ represent a neural network with $p$ weights; in contrast to the previous section, here $G$ is a function of the \emph{network weights} $\vw \in \reals^p$ with a fixed input $\bz$ (typically chosen at random and then fixed thereafter).  This is the opposite of the previous section, where we treated $\vw$ as fixed (pre-learned) and $\bz$ as varying.  The function $G(\vw)$ is our untrained neural network, and the fixed input $\bz$ is considered part of the network.

The architecture of the network is critical, and is discussed in more detail later. For now, we note that a good choice for images is a simple five-layer convolutional network. 
We reconstruct an image by applying an optimization procedure (typically gradient descent) starting from a random initialization of the network weights, using the least-squares loss:
\begin{align}
\label{eq:untrainedoptprob}
\mathcal L(\vw) = \norm{ \vy - \mA G(\vw) }^2_2.
\end{align}
This optimization procedure, possibly with early-stopped iterations for regularization, yields the estimate $\hat \vw$, from which we estimate the unknown signal as $\hat \vx = G(\hat \vw)$.

This general approach is based on the empirical observation that untrained convolutional networks tend to fit a single natural image significantly faster than pure noise when optimized with gradient descent.  However, for the method to work well, a good choice of architecture, optimization procedure, and regularization (e.g., early stopping) can be critical. 

{\bf Deep image prior.} 
Ulyanov \emph{et al.}~\cite{ulyanov_DeepImagePrior_2018} first observed that using a standard convolutional auto-encoder (the popular U-net~\cite{ronneberger_UNetConvolutionalNetworks_2015}) as a generator network, and regularizing with early stopping, enables excellent denoising performance.  This method has been termed \emph{deep image prior}.

{\bf Deep decoder.} 
Many elements of auto-encoders turn out to be largely irrelevant to the strong performance of deep image prior.  A more recent paper of Heckel and Hand~\cite{heckel_DeepDecoderConcise_2019} proposed a much simpler network architecture, termed the deep decoder. 
This network can be seen as retaining only the most relevant components of a convolutional autoencoder architecture to function as an image prior, and can be obtained from a standard convolutional autoencoder by removing the encoder, the skip connections, and perhaps most notably, the trainable convolutional filters of spatial extent larger than one.

\subsection{Theory}

Untrained neural networks enable provable denoising and compressive sensing. Here we discuss the associated recovery guarantees, along with intuition on when and why untrained neural networks enable accurate signal reconstruction.

\subsubsection{Under-parametrized untrained neural networks}
We say that an untrained neural network is \emph{under-parametrized} if it has fewer parameters $p$ (i.e., the dimension of $\vw$) than its output dimension $n$, and \emph{over-parametrized} otherwise. 
Untrained networks enable signal reconstruction in both regimes. 
We start with the under-parametrized regime, since it is conceptually simpler. 

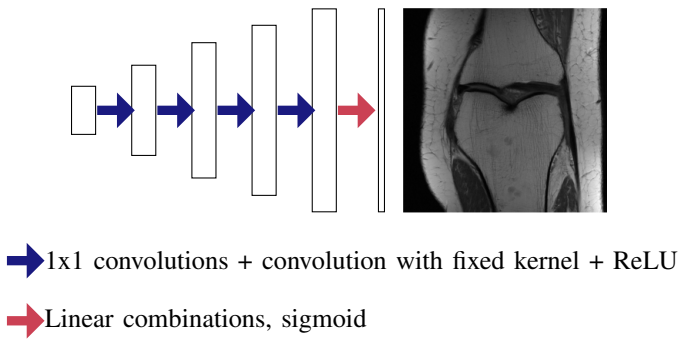
\begin{figure}
\begin{centering}
\begin{tikzpicture}[scale = 0.8]
\def\w{0.2}

\tikzset{arrow1/.style={
    fill = DarkBlue,
    scale=0.5,
    single arrow head extend=0.35cm,
    single arrow,
    minimum height=1cm,
}}

\foreach \x/\y/\label in {1/0.4cm/1,2/0.75cm/2,3/1.125cm/3,4/1.4142,5/1.6875/d}{
\draw (\x-\w, -\y) rectangle (\x+\w,\y);
}

\foreach \x in {1,2,3,4}{
\node at (\x+0.5,0) [arrow1] {};
}

\node at (5+0.5,0) [arrow1,BrickRed] {};
\def\ww{0.1}
\def\xx{6}
\draw (\xx-\ww, -1.6875) rectangle (+\xx,+1.6875);

\node at (8,0) {\includegraphics[width=2.7cm]{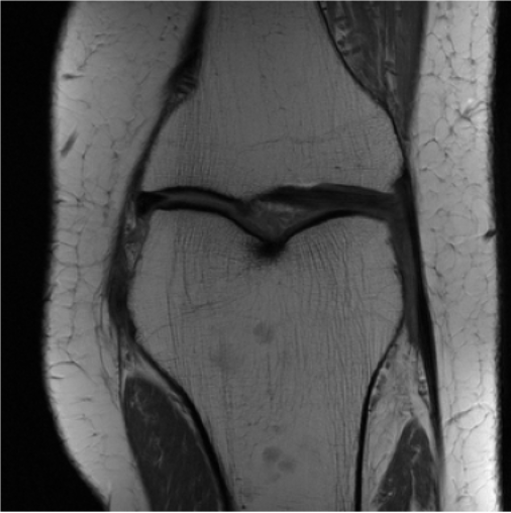}};

\begin{scope}[yshift=-0.3cm,xshift=-1cm]
\node at (1,-2.2) [arrow1] {};
\node at (1.2,-2.2) [right] {
1x1 convolutions + convolution with fixed kernel + ReLU
};

\node at (1,-3.2) [arrow1,BrickRed] {};
\node at (1.2,-3.2) [right] {Linear combinations, sigmoid};
\end{scope}

\end{tikzpicture}
\end{centering}
\caption{
\label{fig:DD}
A rough illustration of the deep decoder, a five-layer untrained convolutional neural network. The network performs 1x1 convolutions (i.e., linear combinations of channels) followed by convolutions with a fixed kernel to map one volume to another. Convolution with a fixed kernel often includes an upsampling operation, as displayed here.
}
\end{figure}

The deep decoder~\cite{heckel_DeepDecoderConcise_2019} is a neural network that transforms a random input volume\footnote{The input vector that we previously denoted by $\bz$ corresponds to the vectorization of $\mB_0$.  Here it is convenient to work with a matrix-valued input.} $\mB_0 \in \reals^{n_0 \times k_0}$ to an output image by applying convolutions with a fixed convolutional upsampling kernel, followed by weighted linear combinations of the channels, followed by an application of ReLU nonlinearities, and repeating these operations several times, e.g., five times for a five-layer network.  See Figure~\ref{fig:DD} for a visualization.  In the simpler case of only two layers, we have $\mB_0 \in \reals^{\frac{n}{2} \times k_0}$, and the deep decoder network is described as follows:
\begin{align}
\label{eq:twolayerDD}
G(\vw) = \relu( \mU_0\mB_0 \mW_0 ) \vw_1,
\end{align}
where $\mU_0 \in \reals^{n\times \frac{n}{2}}$ is a linear operator implementing a convolution with a fixed upsampling operator, and $\mW_0 \in \mathbb R^{k_0\times k_0}$ is a parameter matrix forming linear combinations of the channels $\mU_0\mB_0$. Finally, we apply a ReLU non-linearity and again form linear combinations through multiplication with the parameter vector $\vw_1 \in \reals^{k_0}$, which yields the output image. The parameters of the network are the weights $\vw = (\mW_0,\vw_1)$.

When the number of layers and the number of channels $k_0$ are not too large, this network is a concise image model, in that it can represent a natural image with much fewer network parameters than pixels. 
For example, \cite[Fig.~1]{heckel_DeepDecoderConcise_2019} shows that representing (or compressing) natural images with a deep decoder network that has 30 times fewer parameters than weights gives only a small loss in image quality.  Moreover, for a given storage requirement, the image quality typically surpasses that of sparse wavelet representation, which is the basis for the JPEG2000 compression standard.

To summarize, the deep decoder can represent a natural image with very few parameters. At the same time, in the under-parametrized regime, it cannot represent random noise well; informally, an $n$-dimensional Gaussian noise vector requires roughly $p$ parameters to represent a fraction of $\frac{p}{n}$ of its energy.  For the two-layer deep decoder, this is formalized in the following proposition.  

\begin{proposition}
\label{eq:proppdd}
\emph{(Lack of Noise Fitting with Under-parametrized Networks \cite[Prop.~1]{heckel_DeepDecoderConcise_2019})}
Consider the two-layer deep decoder~\eqref{eq:twolayerDD} with $p$ parameters and arbitrary upsampling and input matrices. Let $\veta$ be zero-mean Gaussian noise with identity covariance matrix.  Then, with high probability,
\begin{align}
\min_{\vw} \norm{G(\vw) - \veta}_2^2 \geq \norm{\veta}_2^2 \left( 1 - c \frac{p \log n}{n} \right),
\end{align}
where $c$ is a numerical constant.
\end{proposition}

Here and throughout this section, we state most results with the terminology ``with high probability'' used informally to avoid overly technical statements, but the precise forms can be found in the references given.

Proposition \ref{eq:proppdd} reveals that when fitting an under-parametrized deep decoder to a noisy image (by minimizing the loss in~\eqref{eq:untrainedoptprob}), we expect to fit only a small amount of noise, thus enabling denoising. 
The number of network parameters, $p$, trades off how well the network fits the underlying signal (larger is better) and how much noise it fits (smaller is better).  

Beyond denoising, similar ideas can be applied to compressive sensing.  Specifically, the proof of Proposition~\ref{eq:proppdd} establishes that any signal generated by an under-parametrized deep decoder lies in a union of low-dimensional subspaces.  Hence, taking measurements with sufficiently many i.i.d.~Gaussian measurements guarantees that only one such signal is consistent with the measurements.


\subsubsection{Over-parametrized untrained neural networks}
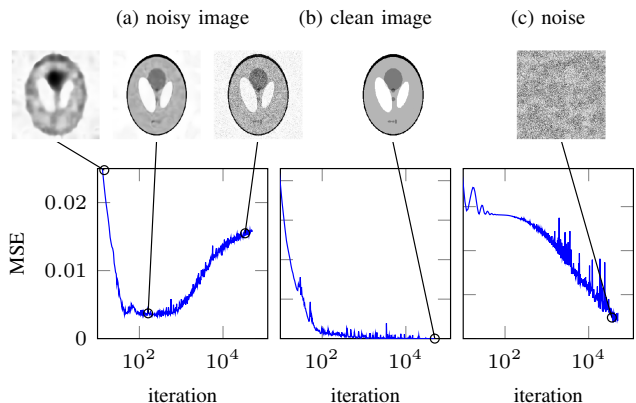
\begin{figure}[t]
\begin{center}
\resizebox{0.48\textwidth}{!}{
\begin{tikzpicture}
\pgfplotsset{every tick label/.append style={font=\scriptsize}}
\begin{groupplot}[
y tick label style={/pgf/number format/.cd,fixed,precision=2},
scaled y ticks = false,
legend style={at={(1,1)}},
         title style={at={(0.5,3.35cm)}, anchor=south},
         group
         style={group size= 3 by 1, xlabels at=edge bottom, ylabels at=edge left,yticklabels at=edge left,
           horizontal sep=0.15cm}, xlabel={\scriptsize iteration}, ylabel={\scriptsize MSE},
         width=3.6cm,height=3.6cm, ymin=0,xmin=10]
	\nextgroupplot[title = {\scriptsize (a) noisy image},xmode=log,ymax=0.025] 
	\addplot +[mark=none] table[x index=0,y index=3]{./figs/DD_denoising_curves.dat};
	

	\nextgroupplot[title = {\scriptsize (b) clean image},xmode=log]
	\addplot +[mark=none] table[x index=0,y index=1]{./figs/DD_denoising_curves.dat};

	\nextgroupplot[title = {\scriptsize(c)  noise},xmode=log]
	\addplot +[mark=none] table[x index=0,y index=2]{./figs/DD_denoising_curves.dat};

\end{groupplot}          

\draw (0.08,2) circle(1.5pt) -- (-0.5cm,2.35cm) node[above,inner sep=-0.5]{
\includegraphics[width=1.1cm]{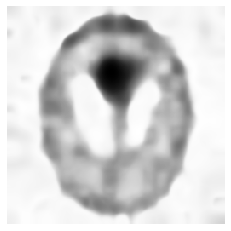}
};

\draw (0.6,0.3) circle(1.5pt) -- (0.7cm,2.35cm) node[above,inner sep=-0.5,draw]{
\includegraphics[width=1.1cm]{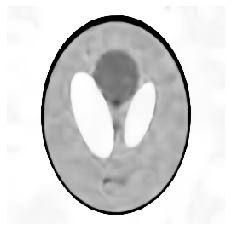}
};

\draw (1.75,1.25) circle(1.5pt) -- (1.9cm,2.35cm) node[above,inner sep=-0.5]{
\includegraphics[width=1.1cm]{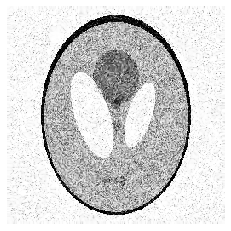}
};


\draw (4,0) circle(1.5pt) -- (3.5cm,2.35cm) node[above,inner sep=-0.5]{
\includegraphics[width=1.1cm]{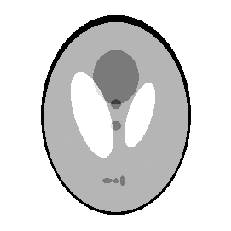}
};

\draw (6.1,0.25) circle(1.5pt) -- (5.5cm,2.35cm) node[above,inner sep=-0.5]{
\includegraphics[width=1.1cm]{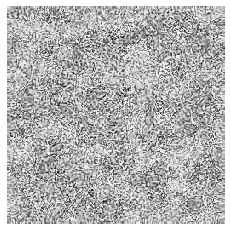}
};
\end{tikzpicture}
}
\end{center}
\vspace{-0.5cm}
\caption{
\label{fig:fittingcurves}
Fitting an over-parametrized deep decoder network to (a) a noisy image, (b) a clean image, and (c) pure noise. Here, MSE denotes Mean Square Error of the network output with respect to the clean image in (a) and fitted images in (b) and (c). While the network can fit the noise due to over-parameterization, it fits natural images with significantly fewer iterations than noise. Hence, when fitting a noisy image, the image component is fitted faster than the noise component, which enables denoising via early stopping. The curves when using other common convolution networks (e.g., a convolutional generator network of a U-net) are very similar.
}
\end{figure}

A sufficiently over-parametrized convolutional neural network can fit any single image perfectly, including noise. Thus, at first sight, it may seem surprising that over-parametrized untrained networks can enable accurate signal reconstruction.  The reason reconstruction is still possible is that, when optimization is performed with gradient descent, the network fits a natural image significantly faster than it fits noise. 
This is illustrated in Figure~\ref{fig:fittingcurves}, where gradient descent is applied to fit a clean image (Fig.~\ref{fig:fittingcurves}(b)) and pure noise (Fig.~\ref{fig:fittingcurves}(c)), by minimizing the least-squares loss~\eqref{eq:untrainedoptprob} with $\mA=\mI$.  After about $300$ iterations the network fits the clean example image, but it requires around $3000$ iterations to fit the noise. If we apply gradient descent with a noisy image (Fig.~\ref{fig:fittingcurves}(a)), the network first fits the image part of the noisy image and only later the noise part. Thus, early stopping at about $300$ iterations denoises the image.

For denoising with an over-parametrized untrained network, regularization via early stopping is critical for performance, since with enough iterations the network fits the entire noisy image. The early stopping time plays an analogous role to the number of parameters for image reconstruction with an under-parametrized neural network: More iterations amounts to fitting the image part better, but also fitting more noise.  

Empirically, untrained neural networks often perform best in the over-parametrized regime. Several variants of convolutional generator networks work well, including the deep decoder, a U-net, and a convolutional generator network~\cite{heckel_DeepDecoderConcise_2019,darestani_AcceleratedMRIUnTrained_2021,ulyanov_DeepImagePrior_2018}. 

{\bf Provable denoising with over-parametrized convolutional networks.} Here we state a theoretical result formalizing the statement that convolutional generators optimized with gradient descent fit natural images faster than noise, and that fitting convolutional generators via early stopped gradient descent provably denoises ``natural'' images. 

\begin{figure}[t]
\begin{center}
\begin{tikzpicture}
\begin{groupplot}[axis lines=none,
y tick label style={/pgf/number format/.cd,fixed,precision=3},
scaled y ticks = false,
legend style={at={(1.2,1)}},
         title style={at={(0.5,-0.7cm)}, anchor=south}, group
         style={group size= 4 by 1, xlabels at=edge bottom,
           horizontal sep=0.3cm, vertical sep=1.3cm},
         width=3.5cm,height=3cm]

\nextgroupplot[title={$\vv_1$}]
\addplot +[mark=none,thick] table[x index=0,y index=1]{./figs/circulant_singular_vectors.dat};
\nextgroupplot[title={$\vv_2$}]
\addplot +[mark=none,thick] table[x index=0,y index=2]{./figs/circulant_singular_vectors.dat};
\nextgroupplot[title={$\vv_6$}]
\addplot +[mark=none,thick] table[x index=0,y index=3]{./figs/circulant_singular_vectors.dat};
\nextgroupplot[title={$\vv_{21}$}]
\addplot +[mark=none,thick] table[x index=0,y index=4]{./figs/circulant_singular_vectors.dat};
\end{groupplot}          
\end{tikzpicture}
\end{center}
\vspace{-0.5cm}
\caption{
\label{fig:circulantsingvectors}
The 1st, 2nd, 6th, and 21st trigonometric basis functions in dimension $n=300$.
}
\end{figure}
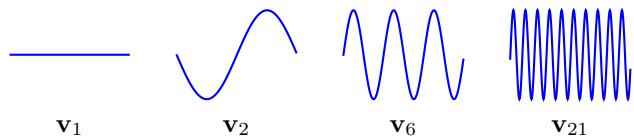

As a suitable model for natural images, we consider smooth signals.  Specifically, a signal $\vx \in \reals^n$ is \emph{$q$-smooth} if it can be represented as a linear combination of the $q$ first trigonometric basis functions (illustrated in Figure~\ref{fig:circulantsingvectors}).  As motivation for this definition, \cite[Fig.~4]{simoncelli_NaturalImageStatistics_2001} shows that the power spectrum (i.e., the energy distribution by frequency) of a natural image decays rapidly from low frequencies to high frequencies.

We consider a randomly initialized network of the form~\eqref{eq:twolayerDD}. 
The result stated below relies on the insight that the behavior of large over-parametrized neural networks is dictated
by the spectral properties of its Jacobian mapping at initialization.\footnote{The Jacobian of the function $G\colon \reals^p \to \reals^n$ is the matrix $\mathbf{J} \in \reals^{n\times p}$ whose $(i,j)$-th entry is equal to the derivative of the $i$-th output value with respect to the $j$-th input value.} 
The left-singular values of the expected Jacobian of this convolutional network at initialization are the trigonometric basis functions $\vv_1,\ldots, \vv_n$. Provided that the fixed convolutional filter of the convolution operation $\mU_0$ in the deep decoder network is relatively narrow (which it is in practice), the associated singular values $\sigma_1 > \sigma_2 > \ldots > \sigma_n$ decay rapidly, so that large singular values are associated with low-frequency trigonometric basis functions and small singular values are associated with high-frequency basis functions. 

The following result shows that for untrained convolutional networks, gradient descent fits the components of the noisy measurement $\vy = \xtrue + \veta$ that align with the trigonometric basis functions at speeds determined by the associated singular values.

\begin{theorem} 
\label{thm:untraineddenoising}
{\em (Denoising Guarantees with Over-parametrized Networks \cite[Thm.~2]{heckel_DenoisingRegularizationExploiting_2020})} 
Assume that $\xtrue$ is a $p$-smooth signal, and let $\veta$ be an arbitrary noise vector. 
Suppose that we fit a randomly initialized network of the form~\eqref{eq:twolayerDD} via gradient descent with step size $\alpha \leq \frac{1}{\sigma_1^2}$ for $t$ iterations to minimize the least-squares loss $\mathcal L(\vw) = \norm{G(\vw) - \by }_2^2$ with $\by = \xtrue + \veta$.  Suppose that the network is sufficiently wide, namely, $k_0 \geq \Omega\big(\frac{n}{\epsilon^4}\big)$, for some $\epsilon>0$. 
Then the estimate of the untrained network based on the $t$-th iterate $\vw_t$ obeys, with high probability over the random initialization,
\begin{align}
\norm{G(\vw_t) - \xtrue}_2
&\leq
(1- \alpha \sigma_p^2)^t \norm{\xtrue}_2 \nonumber \\
&+
\left( \sum_{i=1}^n ( (1-\alpha \sigma_i^2)^t - 1 )^2 \veta^T \vv_i \right)^{1/2} + \epsilon,
\end{align}
where $\{\sigma_i\}_{i=1}^n$ are the singular values described above.
\end{theorem}

In this result, $\epsilon$ is an error term that becomes negligible if the network is sufficiently wide. 
The first term is the error for fitting the signal, and the second term corresponds to the noise fitted after $t$ iterations. 
The signal-fitting term decreases to zero in the number of iterations, while the noise-fitting term increases in the number of iterations, up to the noise energy. Thus, there is a trade-off between signal-fitting and noise-fitting. 
After sufficiently many iterations, $(1- \alpha \sigma_p^2)^t$ is small, and thus so is the signal fitting error. At the same time, after such a number of iterations, only the components of the noise that align with (roughly) the $p$-many lowest frequency trigonometric basis functions are fitted, provided that the singular values decay sufficiently fast. 

A consequence of this result (see \cite[Thm.~1]{heckel_DenoisingRegularizationExploiting_2020}) is that there is an optimal number of iterations such that for denoising a signal corrupted with Gaussian noise $\veta \sim \mathcal \calN(\bzero,\mI)$, the estimate based on early-stopped gradient descent obeys 
\begin{align*}
\norm{G(\vw_t) - \xtrue}_2  \leq O\left( \frac{p}{n}\right).
\end{align*}
This ensures that only a fraction $\frac{p}{n}$ of the noise energy is fitted, and the rest of the noise that lies outside of the signal subspace spanned by the $p$-lowest frequency trigonometric basis functions is filtered out. That is, up to a constant factor, one attains optimal performance for denoising a $p$-smooth signal with Gaussian noise.


{\bf Provable compressive sensing with over-parametrized convolutional networks.}
Here we consider signal reconstruction from $m \ll n$ noiseless random Gaussian measurements with an untrained network, using gradient descent applied to the loss~\eqref{eq:untrainedoptprob}. For this setup, perhaps surprisingly, no regularization is necessary (in contrast to the denoising problem discussed above) since the network has an interesting self-regularization property. 

The following result is a specialized version of that in \cite[Thm.~2]{heckel_CompressiveSensingUntrained_2020}, which considers general decay patterns of the singular values of the Jacobian.  Sufficiently fast decay is needed for accurate reconstruction, and the following result focuses on geometric decay, which is motivated by the fast decay typically observed in practice.

\begin{theorem}
\label{thm:untrainedcs}
\emph{(Compressive Sensing Guarantees with Over-parametrized Networks; Corollary of \cite[Theorem 2]{heckel_CompressiveSensingUntrained_2020})}
Let $\mA \in \reals^{m\times n}$ be an i.i.d.~Gaussian random matrix, and suppose that we are given noiseless measurements $\vy = \mA \xtrue$ of an $\frac{m}{3}$-smooth signal $\xtrue$. Consider the two-layer neural network~\eqref{eq:untrainedoptprob}, with the convolutional kernel (of the convolution operator $\mU_0$) chosen so that the singular values of the Jacobian of the network at initialization decay geometrically, i.e., $\sigma_i^2 = \gamma^i$ for some $\gamma \in (0,1)$.  Moreover, suppose the network is sufficiently wide, namely, the number of channels satisfies $k_0 \geq C\frac{m}{\xi^8}$ for some $\xi \in (0,1)$ and a numerical constant $C$. Then, with high probability, the estimate $\vw_\infty$ obtained by applying gradient descent to the loss~\eqref{eq:untrainedoptprob} until convergence satisfies
\begin{align}
\norm{G(\vw_\infty) - \xtrue}_2^2 \leq O\left( \frac{\gamma^{m/3}}{1-\gamma} \norm{\vx}_2^2 \right) + \xi^2 \|\xtrue\|_2^2.
\end{align}
\end{theorem}

This result guarantees the almost perfect recovery of an $\frac{m}{3}$-smooth signal from only $m$ noiseless measurements, which is optimal up to a constant.  Note that the guarantee is non-uniform, i.e., it holds with high probability over $\bA$ for fixed $\xtrue$.  The mechanism underlying this result is that gradient descent fits the lowest-frequency components of the signal before the higher frequency component, similar to the denoising result stated in Theorem \ref{thm:untraineddenoising}. 


\subsection{Discussion and Ongoing Challenges}

{\bf Linear approximation and its limitations.} The proofs of Theorems \ref{thm:untraineddenoising} and \ref{thm:untrainedcs} rely on relating the dynamics of gradient descent applied to fitting an over-parametrized network to that of gradient descent of an associated linear network. This proof technique has been used in a variety of recent works~\cite{soltanolkotabi_TheoreticalInsightsOptimization_2019,du_GradientDescentProvably_2018b,ronen_ConvergenceRateNeural_2019,jacot_NeuralTangentKernel_2018}. 
As such, the analysis readily extends to deeper neural networks.

However, the main shortcoming of the analysis is that it is constrained to networks operating in a regime where it behaves similar to an associated linear model (implicitly entering via the large-width assumption).  
This is a reasonable first-order approximation of what untrained networks actually do, but in practice untrained networks typically do not operate in the regime where they behave like associated linear models. What makes untrained networks work so well compared to linear models for denoising and signal reconstruction cannot be captured by this analysis, and therefore, an important avenue for future research is to develop a finer analysis for lower-width untrained networks. 

{\bf Beyond convolutional networks.}
In this section, we focused on image reconstruction with untrained convolutional neural networks. We end this section by discussing architectures beyond convolutional networks, and signals beyond images, for which untrained neural networks can still serve as a powerful approach.

For example, coordinate-based neural representations for images, 3D shapes, and other signals have recently emerged as an alternative for traditional discrete representations such as sparse representations or convolutional neural networks. 
They have been employed for surface reconstruction~\cite{williams_DeepGeometricPrior_2019}, representing scenes and view synthesis~\cite{mildenhall_NeRFRepresentingScenes_2020}, and for representing and working with images. 
Such networks are untrained neural networks, as they perform reconstruction without any training, in a similar fashion to convolutional networks.

A key component of many of the coordinate-based neural representations are sinusoidal mappings in the first layer~\cite{mildenhall_NeRFRepresentingScenes_2020,tancik_FourierFeaturesLet_2020a,sitzmann_ImplicitNeuralRepresentations_2020a}. These networks are closely related to convolutional untrained neural networks, since they can be shown to be equivalent to convolutional architectures if sufficiently wide. 

Finally, untrained neural networks have also been used to reconstruct graph signals~\cite{rey_UntrainedGraphNeural_2021}, as well as continuously-indexed objects through fitting probabilistic models~\cite{zhong_ReconstructingContinuousDistributions_2020,rosenbaum_InferringContinuousDistribution_2021}.  We expect that there is significant potential for further theoretical (and practical) developments in these directions.

    \section{Unfolding Methods} \label{sec:unfolding}

Recent years have witnessed a surge of interest in algorithm unfolding (also known as unrolling) techniques to tackle various inverse problems arising in signal processing, image processing, and machine learning \cite{monga2021algorithm}.


Unfolding methods map an iterative solver (algorithm) of an inverse problem onto a recurrent neural network structure. The different iterations of the iterative algorithm correspond to different layers of the neural network structure, with layer parameters corresponding to solver parameters.  Instead of fixing the layer parameters, they are optimized in a data-driven manner using learning algorithms, such as empirical risk minimization via stochastic gradient descent, by leveraging a dataset consisting of input-output examples (i.e., training data).  Compared to most standard neural network architectures, unfolding methods directly capture domain knowledge according to the iterative algorithm they are based on, and they often contain considerably fewer parameters.  Empirically, unfolding methods have achieved state-of-the-art performance in a variety of applications of interest, e.g., being featured prominently in the fastMRI competition.\footnote{\url{https://fastmri.org/}}


We will focus our attention on how unfolding techniques apply to the classical sparse recovery problem, in view of the fact that -- in addition to its myriad of applications -- this is where much of the existing theory-oriented work has arisen.  Moreover, sparse recovery is the problem for which algorithm unfolding was originally proposed in the pioneering work of Gregor and LeCun \cite{gregor2010learning}.  We refer the reader to recent review articles that overview how unfolding applies to numerous other inverse problems in various fields \cite{monga2021algorithm,SEMBAI,shlezinger2022model}.

\subsection{The Classical ISTA Algorithm}
We consider the  problem of recovering a sparse vector $\xtrue$ given (noisy) linear measurements of the form $\by=\bA\xtrue+\boldeta$, as outlined in Section \ref{sec:intro}.  A classical iterative algorithm to recover $\xtrue$ is iterative shrinkage thresholding algorithm (ISTA) \cite{daubechies2004iterative}.  ISTA is closely connected to the Lasso method, whose optimization problem we repeat here for convenience: 
\begin{align}
    \min_{\xopt} \| \by - \bA\xopt \|_2^2 + \lambda  \| \xopt \|_1.
\end{align}
The ISTA algorithm is an instance of a more general class of techniques called \emph{proximal gradient methods}, which roughly work by performing gradient steps on one term ($\| \by - \bA\xopt \|_2^2$ for Lasso) and applying a so-called \emph{proximal mapping} that encourages the other term to be small ($\lambda  \| \xopt \|_1$ for Lasso).

More specifically, given an initialization $\bx_{0}$, the ISTA algorithm produces the following iterates indexed by $t$: 
\begin{align}
   \bx_{t+1} = \Psi_{\lambda/\xi} \left( \bx_{t} + \frac{1}{\xi} \cdot \bA^T \left( \by - \bA \bx_t \right) \right), \label{eq-ista}
\end{align}
where $\xi$ is an upper bound on the largest eigenvalue of $\bA^T \bA$, and $\Psi_{\theta}(\bz)$ is the soft-thresholding function that is applied on each element of a vector argument as follows: $\Psi_\theta (x) = {\sf sign} (x) \cdot \max\{0,|x| - \theta\}$.  

The most well-known unfolding method -- Learned Iterative Shrinkage-Thresholding Algorithm (LISTA) \cite{gregor2010learning} -- leverages this approach to solve the sparse recovery problem using a neural network in a data-driven manner, and is described in the following subsection.

\subsection{ The Unfolding Principle: LISTA }

\begin{figure}
    \begin{centering}
        \includegraphics[width=0.7\columnwidth]{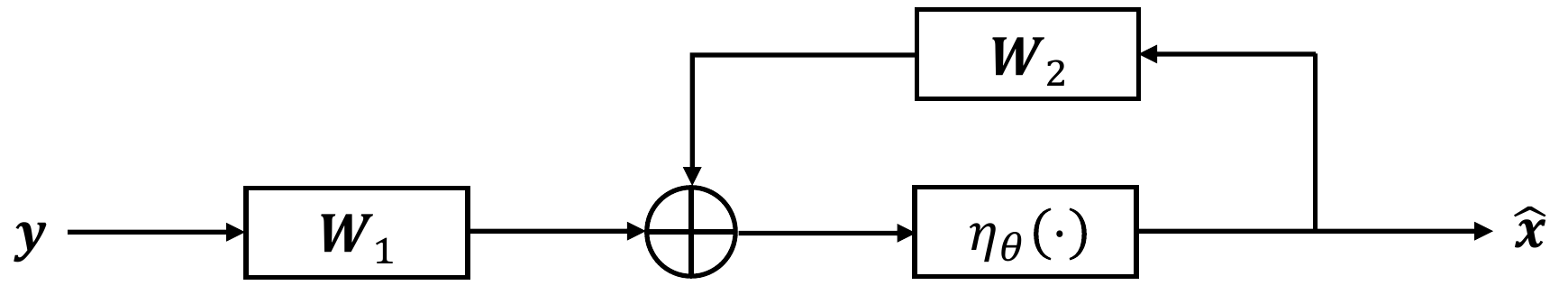}
        \par
        \bigskip
        
        \includegraphics[width=1.0\columnwidth]{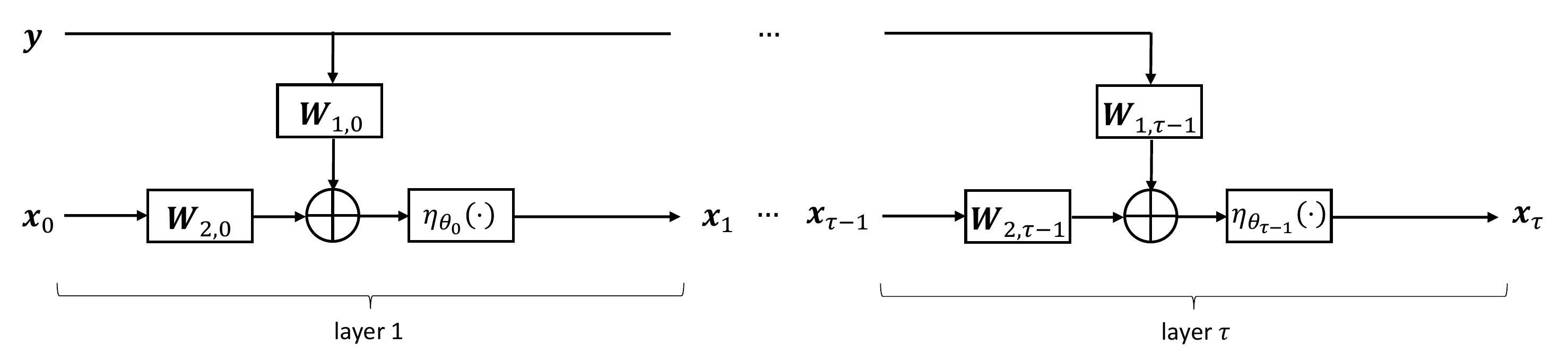}
        \par
    \end{centering}
    \caption{(Top) Recurrent neural network structure defined using a feedback connection. (Bottom) Unrolled feed-forward neural network structure.} 
    \label{unfolding-fcnn}
\end{figure}

The pioneering work of Gregor and LeCun \cite{gregor2010learning} recognized that one can map the iterations of the ISTA algorithm to different layers of a neural network structure. Concretely, by letting $\bW_1 = \frac{1}{\xi} \bA^T$, $\bW_2 = \mathbf{I} - \frac{1}{\xi} \bA^T \bA$, $\theta = \frac{\lambda}{\xi}$ in \eqref{eq-ista}, we can write the $\tau$ iterations of the ISTA algorithm as follows:
\begin{align}
    \bx_{t+1} = \Psi_{\theta} \left(\bW_1 \by + \bW_2 \bx_{t}\right), ~~~~~ t = 0,1,\ldots,\tau-1. \label{eq-unfolding1}
\end{align}
This gives rise to a $\tau$-layer recurrent neural network structure where different layers correspond to different iterations of the ISTA algorithm; see Figure \ref{unfolding-fcnn}. The network non-linearity corresponds to the soft-threshold operator in lieu of the standard ReLU.

Moreover, by letting $\bW_1$, $\bW_2$, and $\theta$ be iteration-dependent (and accordingly denoted by $\bW_{1,t}$, $\bW_{2,t}$, and $\theta_t$), we can write the iterations as follows:
\begin{align}
    \bx_{t+1} = \Psi_{\theta_t} \left(\bW_{1,t} \by + \bW_{2,t} \bx\right), ~~~~~ t = 0,1,\ldots,\tau-1. \label{eq-unfolding2}
\end{align}
This gives rise to a $\tau$-layer feed-forward neural network with side connections, with weight matrices $\bW_{1,t}$ and $\bW_{2,t}$, and non-linearity thresholds $\theta_t$, as illustrated in Figure \ref{unfolding-fcnn}.

In \cite{gregor2010learning}, it was proposed to optimize the parameters of the resulting $\tau$-layer feed-forward neural network in a data-driven manner.  Specifically, given access to a dataset consisting of various (measurement, target) pairs corresponding to the linear model in \eqref{eq:y_linear}, i.e., $\mathcal{S} = \left\{(\by'_i,\bx'_i), i=1,\ldots,N\right\}$ with $\by'_i \in \reals^m$ and $\bx'_i \in \reals^n$, one can consider the following empirical risk minimization problem: 
\begin{align}
    \min_{\bW_{1,t},\bW_{2,t},\theta_t} \frac{1}{N} \sum_{i=1}^{N} \left\| \bx'_i - \hat{\bx}(\by'_i) \right\|_2^2 \label{eq-erm-unfolding}
\end{align}
where $\hat{\bx}(\by'_i)$ is the $\tau$-layer neural network output associated with the network input $\by'_i$.\footnote{Not to be confused with regression and classification problems in which $y$ often denotes the label to be predicted.} This problem can then be solved using optimization techniques such as stochastic gradient descent.  Note that the measurement matrix $\bA$ does not need to be known to apply LISTA, though knowing it can be useful for forming other variations (to be described below).

A common variation is to tie the parameters across the various layers of the network, i.e., $\bW_{1,t}=\bW_1$, $\bW_{2,t}=\bW_2$, and $\theta_t=\theta$. This is particularly helpful when the training set is small. 

The reformulation of $\tau$ ISTA iterations onto a $\tau$-layer neural network with parameters that can be further tuned, as described in \eqref{eq-unfolding1}, \eqref{eq-unfolding2}, and \eqref{eq-erm-unfolding}, is referred to as LISTA. 
It is often referred to as a \textit{model-based learning method}, because the network architecture is specifically defined according to a particular measurement model (the linear model), optimization procedure (Lasso), and iterative solver (ISTA). 
This idea can naturally be extended to many other settings, optimization problems, and solvers.

It has been shown empirically that, in comparison with ISTA, LISTA can deliver a more accurate sparse vector with significantly fewer layers/iterations (e.g., see \cite{gregor2010learning}).  The success of LISTA has spurred numerous applications of algorithm unrolling over the years, in imaging applications \cite{schuler2015learning,jin2017deep,solomon2019deep,bar2021learned,sahel2021deep} and beyond \cite{khobahi2021model,zhang2019real,revach2021kalmannet} (see \cite[Table I]{monga2021algorithm} for a longer list).  More recently, significant interest has arisen in the theoretical foundations of unfolding algorithms.

\subsection{Theoretical Foundations of Unfolding Methods } 

Theoretical studies of unfolding methods broadly fall under the following two categories:

\vspace{0.10cm}

\textit{Optimization/Convergence Results}: This class of results regards convergence properties, studying whether LISTA-type network architectures 
can produce an accurate solution faster compared to ISTA under idealized choices of weights.\footnote{It should be noted, however, that these results typically impose stronger assumptions on the signal, noise, and measurement matrix compared to classical ISTA theory.} This class of contributions also often demonstrates that one can simplify the classical LISTA approach of \cite{gregor2010learning}, e.g., by exploring certain relations/dependencies/couplings between the LISTA learnable parameters.  Works giving results of this kind include \cite{chen2018theoretical,liu2019alista,chen2021hyperparameter}.

\vspace{0.10cm}

\textit{Learning-Theoretic Oriented Results}: Another class of contributions concentrates on learning-theoretic aspects, studying how the generalization error -- corresponding to the difference between the expected error and the empirical error -- behaves as a function of various quantities relating to the learning problem, including the number of training samples. Works giving results of this kind include \cite{behboodi2020compressive,schnoor2021generalization,chen2020understanding,shultzman2022}. 

\vspace{0.10cm}

We proceed by highlighting some key results of both kinds.

\vspace{0.10cm}

\subsubsection{Optimization/Convergence Results} 
In \cite{chen2018theoretical}, Chen \emph{et al.}~showed that the LISTA learnable weight matrices asymptotically admit a partial weight coupling relationship given by 
\begin{align}
    \bW_{2,t} = \mathbf{I} - \bW_{1,t} \bA.
\end{align}
Accordingly, they simplified the LISTA structure as follows:
\begin{align}
    \bx_{t+1} = \Psi_{\theta_t} \left( \bx_t  + \bW_t^T \left( \by - \bA \bx_t \right)\right). \label{lista-cp}
\end{align}
Note that this simplification requires knowledge of the matrix $\bA$, which is not always known in practice. 
This simplified version of LISTA -- which involves learning only a single weight matrix and threshold per layer -- admits the following convergence guarantee.

\begin{theorem} \label{thm:unfold1}
\emph{(Adapted from \cite[Thm.~2]{chen2018theoretical})}
Assume that $\xtrue \in \left\{\bx \in \reals^n: \|\bx\|_0 \leq s, \|\bx\|_{\infty} \leq B\right\}$ and $\|\veta\|_1 \leq \sigma$.  Moreover, assume that $\bA$ satisfies a coherence condition, and that $s$ is sufficiently small as a function of the associated coherence parameter (see \cite[App.~B]{chen2018theoretical} for a formal statement).  Then, there exists a sequence of parameters $\left\{\bW_t,\theta_t\right\}$ such that the sequence of iterates in \eqref{lista-cp} with $\bx_0 = \bzero$ satisfies
\begin{align}
    \left\|\bx_t - \xtrue\right\|_2 \leq s B \exp(-ct) + C \sigma,
\end{align}
where $c>0$ and $C>0$ are scalars that depend only on the linear operator $\bA$ and signal sparsity $s$. 
\end{theorem}

This result shows that a LISTA-like structure can produce a sequence of iterates that is linearly convergent for some sequence of parameters.  In contrast, ISTA is generally sublinearly convergent until its iterates settle on a support \cite{beck2009fast}.  Thus, this result provides theoretical evidence that a LISTA-like structure can outperform conventional methods.\footnote{It has been shown that ISTA can exhibit faster convergence rates provided that one can choose the Lasso regularization parameter $\lambda$ adaptively over iterations \cite{hale2008fixed,xiao2013proximal}. This idea is actually adopted in LISTA, because the parameters $\{\theta_t\}_{t \ge 1}$ correspond to a path of Lasso parameters $\{\lambda_t\}_{t \ge 1}$.}

It should be noted that the sequences of parameters shown to exist in Theorem \ref{thm:unfold1} do not necessarily correspond to the parameters learned using empirical risk minimization.  Thus, results of this kind serve as a \emph{justification for the architecture}, rather than a justification of the training procedure.  Proving analogous results for empirical risk minimization would be of significant interest in future work.

Another variation considered in \cite{chen2018theoretical} is
\begin{align}
    \bx_{t+1} = \Psi_{p_t,\theta_t}^{\rm ss} \left( \bx_t  + \bW_t^T \left( \by - \bA \bx_t \right)\right), \label{lista-cpss}
\end{align}
where one replaces the original soft-thresholding operator $\Psi_{\theta_t} (\cdot)$ with a thresholding operator with support selection $\Psi_{p_t,\theta_t}^{\rm ss} (\cdot)$. This operator retains a proportion $p_t$ of the entries as the ``trusted support'' at layer $t$, where $p_t$ is a hyper-parameter that is manually tuned.  Specifically, it is proposed to choose $p_t$ proportional to $t$ and capped to a maximal value: 
\begin{equation}
    p_t = \min\{p t, p_{\max}\}, \label{eq:pt}
\end{equation}
leaving only $p$ and $p_{\max}$ to be tuned.  This LISTA-like architecture with support selection can exhibit a convergence guarantee that is slightly better than that of Theorem \ref{thm:unfold1}, as stated in the following. 

\begin{theorem} \label{thm:unfold2}
\emph{(Adapted from \cite[Thm.~3]{chen2018theoretical})} 
Under the conditions of Theorem \ref{thm:unfold1}, there exists a sequence of parameters $\left\{\bW_t,\theta_t\right\}$ such that the sequence of iterates in \eqref{lista-cpss} with $\bx_0 = \bzero$ and $p_t$ in \eqref{eq:pt} satisfies
\begin{align}
    \left\|\bx_t - \bx\right\|_2 \leq s B \exp\Big(-\sum_{i=0}^{t-1}c'_i\Big) + C' \sigma,
\end{align}
where $c'_i \geq c$ ($\forall i$) and $C' \leq C$, with $(c,C)$ coming from Theorem \ref{thm:unfold1}.  Moreover, under an additional assumption that the SNR is not too small \cite[Assump.~2]{chen2018theoretical}, we have the strict inequalities $c'_i > c$ for large enough $i$, and $C' < C$.
\end{theorem}

Recent works have also shown that the LISTA structure can be simplified further, without affecting (or even improving) the convergence rates; we proceed by outlining some examples.

\textbf{Analytic LISTA (ALISTA).} 
In the noiseless setting, it was shown in \cite{liu2019alista} that the LISTA structure in \eqref{lista-cpss} can be simplified further to
\begin{align}
    \bx_{t+1} = \Psi_{p_t,\theta_t}^{\rm ss} \left( \bx_t  + \gamma_t \bW^T \left( \by - \bA \bx_t \right)\right). \label{alista}
\end{align}
The matrix $\bW$ -- which is fixed across different layers -- can be pre-computed by solving a data-free optimization problem (which depends only on $\bA$), whereas the layer-wise threshold parameters $\{\theta_t\}_{t=0}^{\tau-1}$ and the layer-wise step-size parameters $\{\gamma_t\}_{t=0}^{\tau-1}$ are optimized using data. The parameters $\{p_t\}_{t=0}^{\tau-1}$ are again chosen according to \eqref{eq:pt}. 
This scheme, known as analytic LISTA (ALISTA), has considerably fewer parameters to learn/train compared to the scheme in \cite{chen2018theoretical} or conventional LISTA \cite{gregor2010learning}. Moreover, this simplified structure retains the linear convergence properties of the structure in \cite{chen2018theoretical}. Note that this variation requires knowledge of $\bA$.

\textbf{Hyper-LISTA.} In \cite{chen2021hyperparameter}, it was proposed to augment the ALISTA structure in \eqref{alista} with an additional momentum term:
\begin{align}
    \bx_{t+1} = \Psi_{p_t,\theta_t}^{\rm ss} \left( \bx_t  + \gamma_t \bW^T \left( \by - \bA \bx_t \right) + \beta_t (\bx_t - \bx_{t-1})\right). \label{eq:lista_SS}
\end{align}
This structure contains the learnable parameters $p_t$, $\theta_t$, $\gamma_t$, and $\beta_t$, with $\bW$ being pre-computed by solving a data-free optimization problem (again depending only on $\bA$).
With this momentum term, it was shown in
\cite{chen2021hyperparameter} that the resulting network can exhibit a better linear convergence rate. They also prove that with instance-optimal parameters -- where $p_t$, $\theta_t$, $\gamma_t$, and $\beta_t$ depend on $\bx_{t}$ -- the network exhibits super-linear convergence. Importantly, with such instance-optimal parameters, it is shown that the tuning procedure involves learning only three hyper-parameters. This ultra lightweight scheme, known as HyperLISTA, is therefore much simpler than the original LISTA or even ALISTA.

\textbf{Other works.} There have been various other works suggesting how to further improve unfolded ISTA networks and its variants  \cite{ablin2019learning,wu2020sparse,li2021learned}.  For example, \cite{ablin2019learning} studies strategies for LISTA that involve learning only step sizes -- named Step-LISTA (SLISTA) -- and that can outperform standard LISTA.  Other earlier works studying the merit of unfolding methods include \cite{moreau2017understanding,xin2016maximal}, and cover distinct algorithms such as iterative hard thresholding (IHT).

\subsubsection{Learning Results} 

We now overview learning-theoretic oriented results that further illuminate the merits of LISTA networks in comparison to standard neural networks. This class of emerging results expound how the expected (population) error deviates from the empirical error as a function of certain quantities relevant to the learning problem.

Suppose that we have access to a training set $\mathcal{S} = \left\{ (\by'_i,\bx'_i) \right\}_{i=1}^N$ containing a series of input-output i.i.d.~samples of (measurement,target) pairs associated with the model in \eqref{eq:y_linear}. This training set is used to learn the learnable parameters of the model-based network (e.g., LISTA) using a learning algorithm such as empirical risk minimization.

We define the population error and the empirical error associated with a certain model-based network $h$ taken from a class of model-based networks $\mathcal{H}$ as
\begin{align}
    \mathcal{L}_{\rm P} (h) = \mathbb{E} \big[ \ell (h; (\by,\bx)) \big],
    \quad \mathcal{L}_{\rm E} (h) = \frac{1}{N} \sum_{i=1}^{N}
    \ell (h; (\by'_i,\bx'_i)),
\end{align}
where $\ell (\cdot ; \cdot)$ represents a per-sample loss function, taken here to be the 
$\ell_2$-loss between the model-based network output for a given input and the associated ground truth. 
The generalization error is defined as follows:
\begin{align}
    {\rm Gen}(h) = \left| \mathcal{L}_{\rm P} (h) - \mathcal{L}_{\rm E} (h) \right|,
\end{align}
which quantifies how much the expected error deviates from the empirical error for a certain model $h \in \mathcal{H}$.

The behaviour of the generalization error for a certain class of model-based networks is discussed by Behboodi \emph{et al.}~\cite{behboodi2020compressive}. Their structure differs slightly from the classical LISTA structure and its variations discussed above.  Specifically, in view of the fact that \cite{behboodi2020compressive} assumes that the signal of interest is sparse in some orthogonal dictionary rather than being sparse itself (i.e., they write the vector of interest $\bx$ in terms of a sparse vector $\bz$ as $\bx = \mathbf{\Phi} \bz$ for some orthogonal dictionary $\mathbf{\Phi} \in \reals^{n \times n}$), their network structure is composed of a LISTA-like multi-layer encoder that converts the measurement vector onto a sparse vector, followed by a linear decoder that converts the sparse vector onto the vector of interest.  For technical reasons, the final output may also be further scaled to have a bounded norm.   Their network structure is also defined  by various weight matrices -- akin to LISTA -- that depend on the forward operator, the dictionary, and other quantities, but the trainable parameters correspond only to the dictionary entries, and are tied across layers. 
(i.e., the dictionary parameterizes this class of model-based networks). Their approach therefore also
requires knowledge of $\bA$. 

\begin{theorem} \label{thm:gen_bound}
{\em (Adapted from \cite[Thm.~2]{behboodi2020compressive})} The generalization error associated with the above-described $\tau$-layer model-based network behaves as follows with high probability:\footnote{For example, with probability 0.99 or any other fixed value in $(0,1)$ (the precise value only affects the hidden constant in the $O(\cdot)$ notation).}
\begin{align}
    {\rm Gen}(h) \le O\bigg( \sqrt{\frac{m n \log{\tau} + n^2 \log{\tau}}{N}} \bigg), \label{eq:gen_bound}
\end{align}
provided that $\bA$ has a bounded spectral norm and $\|\xtrue\|_2$ is bounded (see \cite{behboodi2020compressive} for the precise conditions).
\end{theorem}

This result, whose proof relies on a Rademacher complexity analysis, suggests that model-based networks may exhibit better generalization capabilities than traditional neural networks, in line with empirical results \cite{gregor2010learning}. Concretely, this generalization error bound for LISTA-like model-based networks depends on the number of layers only logarithmically, whereas generalization error bounds for traditional neural networks (albeit in classification settings) can scale exponentially in the number of layers \cite{behboodi2020compressive,schnoor2021generalization}.  On the other hand, the dependence on $m$ and $n$ in \eqref{eq:gen_bound} remains fairly strong, and it would be of interest to determine if it can be reduced, e.g., by exploiting sparsity (notice that the sparsity level $s$ is absent in \eqref{eq:gen_bound}).

\textbf{Other works.} Extensions of the above generalization results are covered in \cite{behboodi2020compressive,schnoor2021generalization,shultzman2022}, involving different learnable parameters or different degrees of weight-sharing between different layers and a variety of network architectures. Notably, \cite{shultzman2022} offers a Rademacher complexity and local Rademacher complexity analysis of the generalization error and estimation error of model-based networks, respectively, showing that the soft-thresholding nonlinearity can play a key role in guaranteeing that model-based networks perform better than traditional neural networks. 
They also show that with a proper choice of parameters, the generalization error bound decays as a function of the number of layers. This result does not hold for standard ReLU networks, demonstrating the power of model-based networks.  
In \cite{pu2022optimization}, further guarantees were given for model-based networks, by deriving a bound on the number of training samples needed to ensure that the training loss decreases to zero as the number of training iterations increases. 

\subsection{Discussion and Ongoing Challenges} 

We conclude this section by discussing some limitations of the existing theory, and the associated ongoing challenges.

\textbf{Convergence rates and training.} As we already highlighted, results such as Theorems \ref{thm:unfold1} and \ref{thm:unfold2} demonstrate the existence of good weights for a given architecture, but it remains an important open challenge to theoretically determine how effective training procedures are in finding good weights, or whether they have provable limitations.  Moreover, previous results in this line of works often impose somewhat restrictive assumptions (e.g., coherence properties of $\bA$ and low sparsity) that it would be of interest to relax or remove.

\textbf{Improved generalization analyses.} Overall, the generalization properties of model-based networks deriving from unrolling techniques are still in their infancy. 
We highlighted some recent results showing that such objects can, in principle, generalize better than classical neural networks. However, a more complete picture may require a significantly better understanding of both model-based networks and standard networks.  For instance, in over-parametrized settings, existing theory may suggest overfitting, but in practice the network may still generalize exceptionally well.  It is of interest to develop new theoretical machinery that captures the interplay between key elements of the learning problem, including the influence of the optimization procedure. Some initial results exploring the interplay between algorithmic notions (e.g., convergence, stability, and sensitivity) and statistical notions (e.g. generalization) appear in \cite{chen2020understanding} within the context of deep architectures with (unrolled) reasoning layers.

\textbf{Beyond sparse recovery.} We have focused on model-based networks for sparse recovery problems, deriving from a Lasso formulation and an associated ISTA solver. However, one can also derive model-based networks for numerous other inverse problems and information processing tasks.  Thus, there remains considerable room for expanding the scope of the existing theory and algorithms, and understanding how model-based networks compare to classical methods or standard neural network architectures.

    \section{Other Topics on Deep Learning Methods in Inverse Problems} \label{sec:other}

In this section, we briefly highlight some other topics that have been considered regarding deep learning methods in inverse problems (without seeking to be exhaustive), including certain areas where theory is largely or completely lacking.

\textbf{Plug-and-play methods.} While denoising is a seemingly relatively simple inverse problem, powerful strategies have been devised for using denoising as a building block for considerably more general inverse problems, e.g., \cite{venkatakrishnan2013plug,metzler2016denoising,romano2017little}.  As an example, the pioneering work \cite{venkatakrishnan2013plug} interpreted the iterative ADMM algorithm as alternating between an $\ell_2$-regularized recovery problem and a denoising problem, and accordingly proposed to use a generic denoiser for the latter (e.g., a pre-trained neural network based denoiser).  The prior information on the signal is then encoded in the denoiser, and accordingly, this approach was termed \emph{plug-and-play priors}.  Related ideas have since been used in AMP algorithms \cite{metzler2016denoising} and \emph{regularization by denoising} \cite{romano2017little}, among others.

A variety of theoretical guarantees, particularly optimization convergence guarantees, have been devised for these methods, e.g., see \cite{chan2016plug,meinhardt2017learning,meinhardt2017learning,ryu2019plug,reehorst2018regularization,tirer2018image,sun2021scalable,liu2021recovery} and the references therein.  In particular, we highlight the recent work \cite{liu2021recovery}, which adopted a restricted eigenvalue condition (REC) analogous to the one used to prove Theorems \ref{thm:ub_lipschitz} and \ref{thm:ub_relu}.  Specifically, the REC is defined with respect to the range of the denoiser (rather than the range of a generative model), and it is shown that this leads to accurate estimation of the underlying signal under suitable boundedness and Lipschitz assumptions on the residual function induced by the denoiser.

\textbf{Instabilities in deep learning methods.} In the machine learning literature, it is widely understood that neural networks for classification (and other tasks) can be highly sensitive to adversarial perturbations in the input \cite{szegedy2014intriguing}.  A detailed theoretical and empirical study was recently given around analogous instability issues in inverse problems \cite{gottschling2020troublesome} (following a related empirical study in \cite{antun2020instabilities}); we proceed by highlighting the over-arching idea in this work.

The focus in \cite{gottschling2020troublesome} is on deep learning for the decoder, i.e., training a neural network to map $\by = \bA\bx$ to $\bx$ (or similarly with noise).  Suppose that such a network learns an accurate mapping for two signals $\bx,\bx'$ with outputs $\by,\by'$, and that $\bA(\bx-\bx')$ is small compared to $\bx-\bx'$ itself (i.e., $\bx-\bx'$ is close to the \emph{nullspace} or \emph{kernel} of $\bA$).  This means that we have two (relatively) nearby $\by,\by'$ being mapped to two distant $\bx,\bx'$.  Then, the network becomes \emph{unstable} in the sense that the output is significantly different for two nearby inputs, resulting in sensitivity to adversarial noise.  Perhaps more surprisingly, it is shown in \cite{gottschling2020troublesome} that even sensitivity to well-behaved \emph{random} noise (e.g., Gaussian) can arise from this phenomenon, both in theory and practice.

In some cases, these difficulties could be circumvented by considering a sufficiently well-behaved measurement matrix (e.g., i.i.d.~Gaussian).  However, when one does not have the luxury of being able to design the measurements, the results of \cite{gottschling2020troublesome} point to the idea that learning methods should be \emph{kernel-aware} in the sense of avoiding the above behavior for pairs of signals whose difference is close to the kernel of $\bA$.  Further details and discussions can be found in \cite{gottschling2020troublesome}, and additional results regarding accuracy and stability can be found in \cite{colbrook2022difficulty}.

\textbf{Training, generalization, and out-of-distribution performance.} As we highlighted in Sections \ref{sec:generative_stat} and \ref{sec:unfolding}, theoretical studies of data-driven deep learning methods for inverse problems still largely lack a good understanding of the precise role of training data, including the fundamental notion of generalization.  Beyond the works on unfolding methods highlighted in Section \ref{sec:unfolding}, an example work on the generalization error in inverse problems is \cite{amjad2021deep}, with the generalization bounds depending on (i) a complexity measure of the signal space, and (ii) norms of the Jacobian matrices of both the network itself and the network composed with the forward model.

Moreover, even provably small generalization error on i.i.d.~data may be insufficient in practical scenarios, where one often requires robustness to out-of-distribution samples.  The above-mentioned works on instabilities \cite{gottschling2020troublesome,colbrook2022difficulty} study an important special case of such issues, and another example is mitigating representation error in the case of generative priors \cite{dhar2018modeling,daras2021intermediate}, which we discussed in Section \ref{sec:stat_further}.

Another limitation of the learning-based works that we have surveyed is that they are often based on the availability of ``clean'' training data, e.g., for learning a generative prior or tuning a neural network based decoder.  To address this, various works have explored methods for training with only samples that are noisy (e.g., $\bx+\bz$ instead of $\bx$) or compressed (e.g., $\bA\bx$ instead of $\bx$) \cite{lehtinen2018noise,bora2018ambient,tachella2022unsupervised}.

Overall, despite this initial progress, we believe that much more remains to be done around training and generalization, and that these issues will play a crucial role in future studies of data-driven methods.

\textbf{Measurement matrix design.} Beyond signal modeling and decoding techniques, deep learning methods have been proposed for designing the measurement matrix $\bA$ in compressive sensing \cite{wu2019learning,learn2sample}.  However, these works have largely focused on algorithm design and empirical evaluation, rather than theory.  While theoretical analyses for certain learning-based measurement designs do exist (e.g., \cite{baldassarre2016learning}) with the possibility of specializing to scenarios involving neural networks, the theory of deep learning based measurement design currently appears to remain largely open.

\textbf{Other decoding techniques.} As we outlined in Section \ref{sec:unfolding}, there exist a variety of theoretical results for unfolding algorithms of interest.  However, unfolding methods are just one of many classes of deep learning based decoders \cite{ongie2020deep}, varying according to the architecture, the degree of prior knowledge of $\bA$, and so on.  Accordingly, there remains considerable room for expanding the scope of theoretical studies in this domain.

\textbf{Specialized inverse problems.} Theoretical guarantees for deep learning based inverse problems have largely focused on the important special cases of denoising and compressive sensing, or problems closely related to these.  Further theoretical studies on other specialized inverse problems (e.g., inpainting, super-resolution, etc.) could provide significant benefit to this continually developing research area.   We also highlight the topic of \emph{deep learning for coding and communication} \cite{gunduz2019machine,kim2020physical,shlezinger2019viterbinet,shlezinger2021model2}, which similarly poses a variety of specialized inverse problems whose study has largely relied on empirical evaluation.

    \section{Conclusion}

While studies of deep learning methods are typically driven by their excellent practical performance, they also pose a variety of unique and exciting theoretical questions.  We have surveyed several prominent examples of the theory behind deep learning methods for inverse problems, and outlined a variety of ongoing challenges and open problems.  Overall, despite the rapid growth of this line of works, we believe that the topic remains in its early stages, with many of the most exciting developments still to come.
    
    \section*{Acknowledgments}
    
    J.~Scarlett is supported by the Singapore National Research Foundation (NRF) under grant number R-252-000-A74-281. 
    
    R.~Heckel acknowledges support by the Institute of Advanced Studies at the Technical University of Munich and the Deutsche Forschungsgemeinschaft (DFG, German Research Foundation) - 456465471, 464123524.
    
    P.~Hand is supported by National Science Foundation (NSF) awards DMS-1848087, DMS-2022205, and DMS-2053448.  
    
    Y.~C.~Eldar and M.~R.~D.~Rodrigues are supported by The Weizmann-UK Making Connections Programme (Ref. 129589).


    \bibliographystyle{myIEEEtran}
   \bibliography{refs}

\begin{thebibliography}{100}
\providecommand{\url}[1]{#1}
\csname url@samestyle\endcsname
\providecommand{\newblock}{\relax}
\providecommand{\bibinfo}[2]{#2}
\providecommand{\BIBentrySTDinterwordspacing}{\spaceskip=0pt\relax}
\providecommand{\BIBentryALTinterwordstretchfactor}{4}
\providecommand{\BIBentryALTinterwordspacing}{\spaceskip=\fontdimen2\font plus
\BIBentryALTinterwordstretchfactor\fontdimen3\font minus
  \fontdimen4\font\relax}
\providecommand{\BIBforeignlanguage}[2]{{%
\expandafter\ifx\csname l@#1\endcsname\relax
\typeout{** WARNING: IEEEtranS.bst: No hyphenation pattern has been}%
\typeout{** loaded for the language `#1'. Using the pattern for}%
\typeout{** the default language instead.}%
\else
\language=\csname l@#1\endcsname
\fi
#2}}
\providecommand{\BIBdecl}{\relax}
\BIBdecl

\bibitem{aberdam2020and}
A.~Aberdam, D.~Simon, and M.~Elad, ``When and how can deep generative models be
  inverted?'' \emph{arXiv preprint arXiv:2006.15555}, 2020.

\bibitem{ablin2019learning}
P.~Ablin, T.~Moreau, M.~Massias, and A.~Gramfort, ``Learning step sizes for
  unfolded sparse coding,'' \emph{Conference on Neural Information Processing
  Systems}, 2019.

\bibitem{amjad2021deep}
J.~Amjad, Z.~Lyu, and M.~R.~D. Rodrigues, ``Deep learning model-aware
  regularization with applications to inverse problems,'' \emph{IEEE
  Transactions on Signal Processing}, vol.~69, pp. 6371--6385, 2021.

\bibitem{antun2020instabilities}
V.~Antun, F.~Renna, C.~Poon, B.~Adcock, and A.~C. Hansen, ``On instabilities of
  deep learning in image reconstruction and the potential costs of ai,''
  \emph{Proceedings of the National Academy of Sciences}, vol. 117, no.~48, pp.
  30\,088--30\,095, 2020.

\bibitem{asim2020invertible}
M.~Asim, M.~Daniels, O.~Leong, A.~Ahmed, and P.~Hand, ``Invertible generative
  models for inverse problems: Mitigating representation error and dataset
  bias,'' in \emph{International Conference on Machine Learning}, 2020.

\bibitem{aubin2019precise}
B.~Aubin, B.~Loureiro, A.~Baker, F.~Krzakala, and L.~Zdeborov{\'a}, ``Precise
  asymptotics for phase retrieval and compressed sensing with random generative
  priors,'' in \emph{NeurIPS Workshop on Deep Learning and Inverse Problems},
  2019.

\bibitem{aubin2019spiked}
B.~Aubin, B.~Loureiro, A.~Maillard \emph{et~al.}, ``The spiked matrix model
  with generative priors,'' \emph{Conference on Neural Information Processing
  Systems}, 2019.

\bibitem{ba2010lower}
K.~D. Ba, P.~Indyk, E.~Price, and D.~P. Woodruff, ``Lower bounds for sparse
  recovery,'' in \emph{ACM-SIAM Symposium on Discrete Algorithms}, 2010.

\bibitem{baldassarre2016learning}
L.~Baldassarre, Y.-H. Li, J.~Scarlett, B.~G{\"o}zc{\"u}, I.~Bogunovic, and
  V.~Cevher, ``Learning-based compressive subsampling,'' \emph{IEEE Journal of
  Selected Topics in Signal Processing}, vol.~10, no.~4, pp. 809--822, 2016.

\bibitem{bar2021learned}
O.~Bar-Shira, A.~Grubstein, Y.~Rapson, D.~Suhami, E.~Atar, K.~Peri-Hanania,
  R.~Rosen, and Y.~C. Eldar, ``Learned super resolution ultrasound for improved
  breast lesion characterization,'' in \emph{International Conference on
  Medical Image Computing and Computer-Assisted Intervention}.\hskip 1em plus
  0.5em minus 0.4em\relax Springer, 2021, pp. 109--118.

\bibitem{baraniuk2010model}
R.~G. Baraniuk, V.~Cevher, M.~F. Duarte, and C.~Hegde, ``Model-based
  compressive sensing,'' \emph{IEEE Transactions on information theory},
  vol.~56, no.~4, pp. 1982--2001, 2010.

\bibitem{baraniuk2009random}
R.~G. Baraniuk and M.~B. Wakin, ``Random projections of smooth manifolds,''
  \emph{Foundations of Computational Mathematics}, vol.~9, no.~1, pp. 51--77,
  2009.

\bibitem{beck2009fast}
A.~Beck and M.~Teboulle, ``A fast iterative shrinkage-thresholding algorithm
  for linear inverse problems,'' \emph{SIAM Journal on Imaging Sciences},
  vol.~2, no.~1, pp. 183--202, 2009.

\bibitem{behboodi2020compressive}
A.~Behboodi, H.~Rauhut, and E.~Schnoor, ``Compressive sensing and neural
  networks from a statistical learning perspective,'' \emph{arXiv preprint
  arXiv:2010.15658}, 2020.

\bibitem{berk2023coherence}
A.~Berk, S.~Brugiapagli, B.~Joshi, Y.~Plan, M.~Scott, and O.~Yilmaz, ``A
  coherence parameter characterizing generative compressed sensing with
  {F}ourier measurements,'' \emph{IEEE Journal on Selected Areas in Information
  Theory (to appear)}, 2023.

\bibitem{bickel2009simultaneous}
P.~J. Bickel, Y.~Ritov, and A.~B. Tsybakov, ``Simultaneous analysis of {L}asso
  and {D}antzig selector,'' \emph{The Annals of Statistics}, vol.~37, no.~4,
  pp. 1705--1732, 2009.

\bibitem{bora2017compressed}
A.~Bora, A.~Jalal, E.~Price, and A.~G. Dimakis, ``Compressed sensing using
  generative models,'' in \emph{International Conference on Machine Learning},
  2017, pp. 537--546.

\bibitem{bora2018ambient}
A.~Bora, E.~Price, and A.~G. Dimakis, ``{AmbientGAN}: Generative models from
  lossy measurements,'' in \emph{International Conference on Learning
  Representations}, 2018.

\bibitem{bostan_DeepPhaseDecoder_2020}
E.~Bostan, R.~Heckel, M.~Chen, M.~Kellman, and L.~Waller, ``Deep phase decoder:
  Self-calibrating phase microscopy with an untrained deep neural network,''
  \emph{Optica}, 2020.

\bibitem{candes2005decoding}
E.~Candes and T.~Tao, ``Decoding by linear programming,'' \emph{IEEE
  Transactions on Information Theory}, vol.~51, no.~12, pp. 4203--4215, 2005.

\bibitem{candes2013how}
E.~J. Candes and M.~A. Davenport, ``How well can we estimate a sparse vector?''
  \emph{Applied and Computational Harmonic Analysis}, vol.~34, no.~2, pp.
  317--323, 2013.

\bibitem{candes2006stable}
E.~J. Candes, J.~K. Romberg, and T.~Tao, ``Stable signal recovery from
  incomplete and inaccurate measurements,'' \emph{Communications on Pure and
  Applied Mathematics}, vol.~59, no.~8, pp. 1207--1223, 2006.

\bibitem{chan2016plug}
S.~H. Chan, X.~Wang, and O.~A. Elgendy, ``Plug-and-play {ADMM} for image
  restoration: Fixed-point convergence and applications,'' \emph{IEEE
  Transactions on Computational Imaging}, vol.~3, no.~1, pp. 84--98, 2016.

\bibitem{chen2018theoretical}
X.~Chen, J.~Liu, Z.~Wang, and W.~Yin, ``Theoretical linear convergence of
  unfolded {ISTA} and its practical weights and thresholds,'' in
  \emph{Conference on Neural Information Processing Systems}, vol.~31, 2018.

\bibitem{chen2021hyperparameter}
X.~Chen, J.~Liu, Z.~Wang, and W.~Yin, ``Hyperparameter tuning is all you need
  for {LISTA},'' in \emph{Conference on Neural Information Processing Systems},
  2021.

\bibitem{chen2020understanding}
X.~Chen, Y.~Zhang, C.~Reisinger, and L.~Song, ``Understanding deep architecture
  with reasoning layer,'' in \emph{Conference on Neural Information Processing
  Systems}, vol.~33, 2020, pp. 1240--1252.

\bibitem{clason2017nonsmooth}
C.~Clason, ``Nonsmooth analysis and optimization,'' \emph{arXiv preprint
  arXiv:1708.04180}, 2017.

\bibitem{cocola2022signal}
J.~Cocola, ``Signal recovery with non-expansive generative network priors,''
  \emph{arXiv preprint arXiv:2204.13599}, 2022.

\bibitem{cohen2009compressed}
A.~Cohen, W.~Dahmen, and R.~DeVore, ``Compressed sensing and best $k$-term
  approximation,'' \emph{Journal of the American Mathematical Society},
  vol.~22, no.~1, pp. 211--231, 2009.

\bibitem{colbrook2022difficulty}
M.~J. Colbrook, V.~Antun, and A.~C. Hansen, ``The difficulty of computing
  stable and accurate neural networks: On the barriers of deep learning and
  {S}male’s 18th problem,'' \emph{Proceedings of the National Academy of
  Sciences}, vol. 119, no.~12, p. e2107151119, 2022.

\bibitem{daras2022score}
G.~Daras, Y.~Dagan, A.~Dimakis, and C.~Daskalakis, ``Score-guided intermediate
  level optimization: Fast {L}angevin mixing for inverse problems,'' in
  \emph{International Conference on Machine Learning}, 2022.

\bibitem{daras2021intermediate}
G.~Daras, J.~Dean, A.~Jalal, and A.~Dimakis, ``Intermediate layer optimization
  for inverse problems using deep generative models,'' in \emph{International
  Conference on Machine Learning}, 2021.

\bibitem{darestani_AcceleratedMRIUnTrained_2021}
M.~Z. Darestani and R.~Heckel, ``Accelerated {MRI} with un-trained neural
  networks,'' \emph{IEEE Transactions on Computational Imaging}, vol.~7, pp.
  724--733, 2021.

\bibitem{daskalakis2020constant}
C.~Daskalakis, D.~Rohatgi, and E.~Zampetakis, ``Constant-expansion suffices for
  compressed sensing with generative priors,'' in \emph{Conference on Neural
  Information Processing Systems}, 2020.

\bibitem{daubechies2004iterative}
I.~Daubechies, M.~Defrise, and C.~De~Mol, ``An iterative thresholding algorithm
  for linear inverse problems with a sparsity constraint,''
  \emph{Communications on Pure and Applied Mathematics}, vol.~57, no.~11, pp.
  1413--1457, 2004.

\bibitem{DDEK11}
M.~Davenport, M.~Duarte, Y.~C. Eldar, and G.~Kutyniok, \emph{Compressed
  Sensing: Theory and Applications}.\hskip 1em plus 0.5em minus 0.4em\relax
  Cambridge University Press, 2011, ch. Introduction to Compressed Sensing.

\bibitem{dhar2018modeling}
M.~Dhar, A.~Grover, and S.~Ermon, ``Modeling sparse deviations for compressed
  sensing using generative models,'' in \emph{International Conference on
  Machine Learning}, 2018.

\bibitem{dong2015image}
C.~Dong, C.~C. Loy, K.~He, and X.~Tang, ``Image super-resolution using deep
  convolutional networks,'' \emph{IEEE Transactions on Pattern Analysis and
  Machine Intelligence}, vol.~38, no.~2, pp. 295--307, 2015.

\bibitem{du_GradientDescentProvably_2018b}
S.~S. Du, X.~Zhai, B.~Poczos, and A.~Singh, ``Gradient descent provably
  optimizes over-parameterized neural networks,'' in \emph{International
  {{Conference}} on {{Learning Representations}}}, 2018.

\bibitem{EM09a}
Y.~C. Eldar and M.~Mishali, ``{Robust recovery of signals from a structured
  union of subspaces},'' \emph{IEEE Transactions on Information Theory},
  vol.~55, no.~11, pp. 5302--5316, 2009.

\bibitem{eldar2015sampling}
Y.~C. Eldar, \emph{Sampling theory: Beyond bandlimited systems}.\hskip 1em plus
  0.5em minus 0.4em\relax Cambridge University Press, 2015.

\bibitem{feng2021unifying}
O.~Y. Feng, R.~Venkataramanan, C.~Rush, and R.~J. Samworth, ``A unifying
  tutorial on approximate message passing,'' \emph{Foundations and Trends in
  Machine Learning}, vol.~15, no.~4, pp. 335--536, 2022.

\bibitem{fletcher2018inference}
A.~K. Fletcher, S.~Rangan, and P.~Schniter, ``Inference in deep networks in
  high dimensions,'' in \emph{IEEE International Symposium on Information
  Theory}, 2018.

\bibitem{foucart2013mathematical}
S.~Foucart and H.~Rauhut, \emph{A Mathematical Introduction to Compressive
  Sensing}.\hskip 1em plus 0.5em minus 0.4em\relax Springer New York, 2013.

\bibitem{gilbert2013ell}
A.~C. Gilbert, H.~Q. Ngo, E.~Porat, A.~Rudra, and M.~J. Strauss,
  ``$\ell_2$/$\ell_2$-foreach sparse recovery with low risk,'' in
  \emph{International Colloquium on Automata, Languages, and Programming},
  2013.

\bibitem{gomez2019fast}
F.~L. G\'omez, A.~Eftekhari, and V.~Cevher, ``Fast and provable {ADMM} for
  learning with generative priors,'' \emph{Conference on Neural Information
  Processing Systems}, 2019.

\bibitem{goodfellow2014generative}
I.~Goodfellow, J.~Pouget-Abadie, M.~Mirza, B.~Xu, D.~Warde-Farley, S.~Ozair,
  A.~Courville, and Y.~Bengio, ``Generative adversarial nets,'' in
  \emph{Conference on Neural Information Processing Systems}, 2014.

\bibitem{gottschling2020troublesome}
N.~M. Gottschling, V.~Antun, B.~Adcock, and A.~C. Hansen, ``The troublesome
  kernel: Why deep learning for inverse problems is typically unstable,''
  \emph{arXiv preprint arXiv:2001.01258}, 2020.

\bibitem{gregor2010learning}
K.~Gregor and Y.~LeCun, ``Learning fast approximations of sparse coding,'' in
  \emph{International Conference on Machine Learning}, 2010.

\bibitem{gunn2022regularized}
S.~Gunn, J.~Cocola, and P.~Hand, ``Regularized training of intermediate layers
  for generative models for inverse problems,'' \emph{arXiv preprint
  arXiv:2203.04382}, 2022.

\bibitem{gunduz2019machine}
D.~Gündüz, P.~de~Kerret, N.~D. Sidiropoulos, D.~Gesbert, C.~R. Murthy, and
  M.~van~der Schaar, ``Machine learning in the air,'' \emph{IEEE Journal on
  Selected Areas in Communications}, vol.~37, no.~10, pp. 2184--2199, 2019.

\bibitem{hale2008fixed}
E.~T. Hale, W.~Yin, and Y.~Zhang, ``Fixed-point continuation for
  $\ell_1$-minimization: Methodology and convergence,'' \emph{SIAM Journal on
  Optimization}, vol.~19, no.~3, pp. 1107--1130, 2008.

\bibitem{hand2018phase}
P.~Hand, O.~Leong, and V.~Voroninski, ``Phase retrieval under a generative
  prior,'' in \emph{Conference on Neural Information Processing Systems}, 2018,
  pp. 9154--9164.

\bibitem{hand2021optimalsample}
P.~Hand, O.~Leong$^*$, and V.~Voroninski, ``Optimal sample complexity of
  subgradient descent for amplitude flow via non-{L}ipschitz matrix
  concentration,'' \emph{Communications in Mathematical Sciences}, vol.~19,
  no.~7, pp. 2035--2047, 2021.

\bibitem{hand2018global}
P.~Hand and V.~Voroninski, ``Global guarantees for enforcing deep generative
  priors by empirical risk,'' in \emph{Conference on Learning Theory}, 2018.

\bibitem{hand2019globalieee}
P.~Hand and V.~Voroninski, ``Global guarantees for enforcing deep generative
  priors by empirical risk,'' \emph{IEEE Transactions on Information Theory},
  vol.~66, no.~1, pp. 401--418, 2019.

\bibitem{heckel_DeepDecoderConcise_2019}
R.~Heckel and P.~Hand, ``Deep decoder: Concise image representations from
  untrained non-convolutional networks,'' in \emph{International Conference on
  Learning Representations}, 2019.

\bibitem{heckel2021rate}
R.~Heckel, W.~Huang, P.~Hand, and V.~Voroninski, ``Rate-optimal denoising with
  deep neural networks,'' \emph{Information and Inference: A Journal of the
  IMA}, vol.~10, no.~4, pp. 1251--1285, 2021.

\bibitem{heckel_CompressiveSensingUntrained_2020}
R.~Heckel and M.~Soltanolkotabi, ``Compressive sensing with un-trained neural
  networks: Gradient descent finds the smoothest approximation,'' in
  \emph{International Conference on Machine Learning}, 2020.

\bibitem{heckel_DenoisingRegularizationExploiting_2020}
R.~Heckel and M.~Soltanolkotabi, ``Denoising and regularization via exploiting
  the structural bias of convolutional generators,'' in \emph{International
  Conference on Learning Representations}, 2020.

\bibitem{hegde2012signal}
C.~Hegde and R.~G. Baraniuk, ``Signal recovery on incoherent manifolds,''
  \emph{IEEE Transactions on Information Theory}, vol.~58, no.~12, pp.
  7204--7214, 2012.

\bibitem{huang2021provably}
W.~Huang, P.~Hand, R.~Heckel, and V.~Voroninski, ``A provably convergent scheme
  for compressive sensing under random generative priors,'' \emph{Journal of
  Fourier Analysis and Applications}, vol.~27, no.~2, pp. 1--34, 2021.

\bibitem{hussein2020image}
S.~A. Hussein, T.~Tirer, and R.~Giryes, ``Image-adaptive {GAN} based
  reconstruction,'' in \emph{AAAI Conference on Artificial Intelligence},
  vol.~34, no.~04, 2020, pp. 3121--3129.

\bibitem{hyder_GenerativeModelsLowDimensional_2020}
R.~Hyder and M.~S. Asif, ``Generative models for low-dimensional video
  representation and reconstruction,'' \emph{IEEE Transactions on Signal
  Processing}, vol.~68, pp. 1688--1701, 2020.

\bibitem{jacot_NeuralTangentKernel_2018}
A.~Jacot, F.~Gabriel, and C.~Hongler, ``Neural tangent kernel: Convergence and
  generalization in neural networks,'' in \emph{Conference on Neural
  Information Processing Systems}, 2018.

\bibitem{jagatap_AlgorithmicGuaranteesInverse_2019}
G.~Jagatap and C.~Hegde, ``Algorithmic guarantees for inverse imaging with
  untrained network priors,'' in \emph{Conference on Neural Information
  Processing Systems}, 2019.

\bibitem{jalal2021instance}
A.~Jalal, S.~Karmalkar, A.~Dimakis, and E.~Price, ``Instance-optimal compressed
  sensing via posterior sampling,'' in \emph{International Conference on
  Machine Learning}, 2021.

\bibitem{jalal2020robust}
A.~Jalal, L.~Liu, A.~G. Dimakis, and C.~Caramanis, ``Robust compressed sensing
  using generative models,'' in \emph{Conference on Neural Information
  Processing Systems}, 2020.

\bibitem{jiao2021just}
Y.~Jiao, D.~Li, M.~Liu, X.~Lu, and Y.~Yang, ``Just least squares: Binary
  compressive sampling with low generative intrinsic dimension,'' \emph{arXiv
  preprint arXiv:2111.14486}, 2021.

\bibitem{jin2017deep}
K.~H. Jin, M.~T. McCann, E.~Froustey, and M.~Unser, ``Deep convolutional neural
  network for inverse problems in imaging,'' \emph{IEEE Transactions on Image
  Processing}, vol.~26, no.~9, pp. 4509--4522, 2017.

\bibitem{johnson1984extensions}
W.~B. Johnson and J.~Lindenstrauss, ``Extensions of {L}ipschitz mappings into a
  {H}ilbert space 26,'' \emph{Contemporary Mathematics}, vol.~26, p.~28, 1984.

\bibitem{joshi2021plugin}
B.~Joshi, X.~Li, Y.~Plan, and O.~Yilmaz, ``{PLUGIn}: A simple algorithm for
  inverting generative models with recovery guarantees,'' in \emph{Conference
  on Neural Information Processing Systems}, vol.~34, 2021.

\bibitem{kamath2020power}
A.~Kamath, E.~Price, and S.~Karmalkar, ``On the power of compressed sensing
  with generative models,'' in \emph{International Conference on Machine
  Learning}, 2020.

\bibitem{khobahi2021model}
S.~Khobahi, N.~Shlezinger, M.~Soltanalian, and Y.~C. Eldar, ``{LoRD-Net}: Low
  resolution detection network for deep low-resolution receivers,'' \emph{IEEE
  Transactions on Signal Processing}, vol.~69, pp. 5651--5664, 2021.

\bibitem{kim2020physical}
H.~Kim, S.~Oh, and P.~Viswanath, ``Physical layer communication via deep
  learning,'' \emph{IEEE Journal on Selected Areas in Information Theory},
  vol.~1, no.~1, pp. 5--18, 2020.

\bibitem{kingma2013auto}
D.~P. Kingma and M.~Welling, ``Auto-encoding variational {B}ayes,'' \emph{arXiv
  preprint arXiv:1312.6114}, 2013.

\bibitem{lehtinen2018noise}
J.~Lehtinen, J.~Munkberg, J.~Hasselgren, S.~Laine, T.~Karras, M.~Aittala, and
  T.~Aila, ``{N}oise2{N}oise: Learning image restoration without clean data,''
  in \emph{International Conference on Machine Learning}, 2018.

\bibitem{lei2019inverting}
Q.~Lei, A.~Jalal, I.~S. Dhillon, and A.~G. Dimakis, ``Inverting deep generative
  models, one layer at a time,'' in \emph{Conference on Neural Information
  Processing Systems}, 2019.

\bibitem{li2021learned}
Z.~Li, K.~Wu, Y.~Guo, and C.~Zhang, ``Learned {ISTA} with error-based
  thresholding for adaptive sparse coding,'' \emph{arXiv preprint
  arXiv:2112.10985}, 2021.

\bibitem{lipton2017precise}
Z.~C. Lipton and S.~Tripathi, ``Precise recovery of latent vectors from
  generative adversarial networks,'' \emph{arXiv preprint arXiv:1702.04782},
  2017.

\bibitem{liu2019alista}
J.~Liu and X.~Chen, ``{ALISTA}: Analytic weights are as good as learned weights
  in {LISTA},'' in \emph{International Conference on Learning Representations},
  2019.

\bibitem{liu2021recovery}
J.~Liu, S.~Asif, B.~Wohlberg, and U.~Kamilov, ``Recovery analysis for
  plug-and-play priors using the restricted eigenvalue condition,'' in
  \emph{Conference on Neural Information Processing Systems}, 2021.

\bibitem{liu2022noniterative}
J.~Liu and Z.~Liu, ``Non-iterative recovery from nonlinear observations using
  generative models,'' in \emph{IEEE/CVF Conference on Computer Vision and
  Pattern Recognition}, 2022.

\bibitem{liu2021towards}
Z.~Liu, S.~Ghosh, and J.~Scarlett, ``Towards sample-optimal compressive phase
  retrieval with sparse and generative priors,'' \emph{Confernece on Neural
  Information Processing Systems}, 2021.

\bibitem{liu2020sample}
Z.~Liu, S.~Gomes, A.~Tiwari, and J.~Scarlett, ``Sample complexity bounds for
  1-bit compressive sensing and binary stable embeddings with generative
  priors,'' in \emph{International Conference on Machine Learning}, 2020.

\bibitem{liu2022projected}
Z.~Liu and J.~Han, ``Projected gradient descent algorithms for solving
  nonlinear inverse problems with generative priors,'' in \emph{International
  Joint Conference on Artificial Intelligence}, 2022.

\bibitem{liu2022generative}
Z.~Liu, J.~Liu, S.~Ghosh, J.~Han, and J.~Scarlett, ``Generative principal
  component analysis,'' in \emph{International Conference on Learning
  Representations}, 2022.

\bibitem{liu2020generalized}
Z.~Liu and J.~Scarlett, ``The generalized {L}asso with nonlinear observations
  and generative priors,'' in \emph{Conference on Neural Information Processing
  Systems}, vol.~33, 2020.

\bibitem{liu2020information}
Z.~Liu and J.~Scarlett, ``Information-theoretic lower bounds for compressive
  sensing with generative models,'' \emph{IEEE Journal on Selected Areas in
  Information Theory}, vol.~1, no.~1, pp. 292--303, 2020.

\bibitem{ma2018invertibility}
F.~Ma, U.~Ayaz, and S.~Karaman, ``Invertibility of convolutional generative
  networks from partial measurements,'' in \emph{Conference on Neural
  Information Processing Systems}, vol.~31, 2018.

\bibitem{mccann2017convolutional}
M.~T. McCann, K.~H. Jin, and M.~Unser, ``Convolutional neural networks for
  inverse problems in imaging: A review,'' \emph{IEEE Signal Processing
  Magazine}, vol.~34, no.~6, pp. 85--95, 2017.

\bibitem{meinhardt2017learning}
T.~Meinhardt, M.~Moller, C.~Hazirbas, and D.~Cremers, ``Learning proximal
  operators: Using denoising networks for regularizing inverse imaging
  problems,'' in \emph{IEEE International Conference on Computer Vision}, 2017,
  pp. 1781--1790.

\bibitem{metzler2017learned}
C.~Metzler, A.~Mousavi, and R.~Baraniuk, ``Learned {D-AMP}: Principled neural
  network based compressive image recovery,'' in \emph{Conference on Neural
  Information Processing Systems}, vol.~30, 2017.

\bibitem{metzler2016denoising}
C.~A. Metzler, A.~Maleki, and R.~G. Baraniuk, ``From denoising to compressed
  sensing,'' \emph{IEEE Transactions on Information Theory}, vol.~62, no.~9,
  pp. 5117--5144, 2016.

\bibitem{mildenhall_NeRFRepresentingScenes_2020}
B.~Mildenhall, P.~P. Srinivasan, M.~Tancik, J.~T. Barron, R.~Ramamoorthi, and
  R.~Ng, ``{{NeRF}}: Representing scenes as neural radiance fields for view
  synthesis,'' in \emph{European Conference on Computer Vision}, 2020.

\bibitem{monga2021algorithm}
V.~Monga, Y.~Li, and Y.~C. Eldar, ``Algorithm unrolling: Interpretable,
  efficient deep learning for signal and image processing,'' \emph{IEEE Signal
  Processing Magazine}, vol.~38, no.~2, pp. 18--44, 2021.

\bibitem{moreau2017understanding}
T.~Moreau and J.~Bruna, ``Understanding trainable sparse coding with matrix
  factorization,'' in \emph{International Conference on Learning
  Representations}, 2017.

\bibitem{learn2sample}
S.~Mulleti, H.~Zhang, and Y.~C. Eldar, ``Learning to sample: Data-driven
  sampling and reconstruction of {FRI} signals,'' \emph{arXiv preprint
  arXiv:2106.14500}, 2021.

\bibitem{naderi2021beyond}
A.~Naderi and Y.~Plan, ``Sparsity-free compressed sensing with generative
  priors as special case,'' \emph{IEEE Journal on Selected Areas in Information
  Theory (to appear)}, 2023.

\bibitem{nguyen2021provable}
T.~V. Nguyen, G.~Jagatap, and C.~Hegde, ``Provable compressed sensing with
  generative priors via {L}angevin dynamics,'' \emph{arXiv preprint
  arXiv:2102.12643}, 2021.

\bibitem{ongie2020deep}
G.~Ongie, A.~Jalal, C.~A. Metzler, R.~G. Baraniuk, A.~G. Dimakis, and
  R.~Willett, ``Deep learning techniques for inverse problems in imaging,''
  \emph{IEEE Journal on Selected Areas in Information Theory}, vol.~1, no.~1,
  pp. 39--56, 2020.

\bibitem{peng2020solving}
P.~Peng, S.~Jalali, and X.~Yuan, ``Solving inverse problems via
  auto-encoders,'' \emph{IEEE Journal on Selected Areas in Information Theory},
  vol.~1, no.~1, pp. 312--323, 2020.

\bibitem{price2011epsilon}
E.~Price and D.~P. Woodruff, ``$(1+\epsilon)$-approximate sparse recovery,'' in
  \emph{IEEE Symposium on Foundations of Computer Science}, 2011.

\bibitem{pu2022optimization}
W.~Pu, Y.~C. Eldar, and M.~R. Rodrigues, ``Optimization guarantees for {ISTA}
  and {ADMM} based unfolded networks,'' in \emph{IEEE International Conference
  on Acoustics, Speech and Signal Processing}, 2022.

\bibitem{qiu2020robust}
S.~Qiu, X.~Wei, and Z.~Yang, ``Robust one-bit recovery via {R}e{LU} generative
  networks: Improved statistical rates and global landscape analysis,'' in
  \emph{International Conference on Machine Learning}, 2020.

\bibitem{raginsky2017non}
M.~Raginsky, A.~Rakhlin, and M.~Telgarsky, ``Non-convex learning via stochastic
  gradient langevin dynamics: A nonasymptotic analysis,'' in \emph{Conference
  on Learning Theory}, 2017.

\bibitem{recht2010guaranteed}
B.~Recht, M.~Fazel, and P.~A. Parrilo, ``Guaranteed minimum-rank solutions of
  linear matrix equations via nuclear norm minimization,'' \emph{SIAM review},
  vol.~52, no.~3, pp. 471--501, 2010.

\bibitem{reehorst2018regularization}
E.~T. Reehorst and P.~Schniter, ``Regularization by denoising: Clarifications
  and new interpretations,'' \emph{IEEE Transactions on Computational Imaging},
  vol.~5, no.~1, pp. 52--67, 2018.

\bibitem{revach2021kalmannet}
G.~Revach, N.~Shlezinger, X.~Ni, A.~L. Escoriza, R.~J. van Sloun, and Y.~C.
  Eldar, ``Kalman{N}et: Neural network aided {K}alman filtering for partially
  known dynamics,'' \emph{IEEE Transactions on Signal Processing}, vol.~70, pp.
  1532--1547, 2022.

\bibitem{rey_UntrainedGraphNeural_2021}
S.~Rey, S.~Segarra, R.~Heckel, and A.~G. Marques, ``Untrained graph neural
  networks for denoising,'' \emph{arXiv preprint arXiv:2109.11700}, 2021.

\bibitem{romano2017little}
Y.~Romano, M.~Elad, and P.~Milanfar, ``The little engine that could:
  Regularization by denoising ({RED}),'' \emph{SIAM Journal on Imaging
  Sciences}, vol.~10, no.~4, pp. 1804--1844, 2017.

\bibitem{ronen_ConvergenceRateNeural_2019}
B.~Ronen, D.~Jacobs, Y.~Kasten, and S.~Kritchman, ``The convergence rate of
  neural networks for learned functions of different frequencies,'' in
  \emph{Conference on Neural Information Processing Systems}, 2019.

\bibitem{ronneberger_UNetConvolutionalNetworks_2015}
O.~Ronneberger, P.~Fischer, and T.~Brox, ``U-{{Net}}: Convolutional networks
  for biomedical image segmentation,'' in \emph{Medical {{Image Computing}} and
  {{Computer-Assisted Intervention}}}, 2015, pp. 234--241.

\bibitem{rosenbaum_InferringContinuousDistribution_2021}
D.~Rosenbaum, M.~Garnelo, M.~Zielinski, C.~Beattie, E.~Clancy, A.~Huber,
  P.~Kohli, A.~W. Senior, J.~Jumper, C.~Doersch, S.~M.~A. Eslami,
  O.~Ronneberger, and J.~Adler, ``Inferring a continuous distribution of atom
  coordinates from cryo-{EM} images using {VAEs},'' \emph{arXiv preprint
  arXiv:2106.14108}, 2021.

\bibitem{ryu2019plug}
E.~Ryu, J.~Liu, S.~Wang, X.~Chen, Z.~Wang, and W.~Yin, ``Plug-and-play methods
  provably converge with properly trained denoisers,'' in \emph{International
  Conference on Machine Learning}, 2019.

\bibitem{sahel2021deep}
Y.~B. Sahel, J.~P. Bryan, B.~Cleary, S.~L. Farhi, and Y.~C. Eldar, ``Deep
  unrolled recovery in sparse biological imaging,'' \emph{IEEE Signal
  Processing Magazine}, vol.~39, no.~2, pp. 45--57, 2022.

\bibitem{schnoor2021generalization}
E.~Schnoor, A.~Behboodi, and H.~Rauhut, ``Generalization error bounds for
  iterative recovery algorithms unfolded as neural networks,'' \emph{arXiv
  preprint arXiv:2112.04364}, 2021.

\bibitem{schuler2015learning}
C.~J. Schuler, M.~Hirsch, S.~Harmeling, and B.~Sch{\"o}lkopf, ``Learning to
  deblur,'' \emph{IEEE Transactions on Pattern Analysis and Machine
  Intelligence}, vol.~38, no.~7, pp. 1439--1451, 2015.

\bibitem{shah2018solving}
V.~Shah and C.~Hegde, ``Solving linear inverse problems using {GAN} priors: An
  algorithm with provable guarantees,'' in \emph{IEEE International Conference
  on Acoustics, Speech, and Signal Processing}, 2018.

\bibitem{SEMBAI}
N.~Shlezinger, J.~Whang, Y.~C. Eldar, and A.~G. Dimakis, ``Model-based deep
  learning,'' \emph{arXiv preprint arXiv:2012.08405}, 2021.

\bibitem{shlezinger2022model}
N.~Shlezinger, Y.~C. Eldar, and S.~P. Boyd, ``Model-based deep learning: On the
  intersection of deep learning and optimization,'' \emph{arXiv preprint
  arXiv:2205.02640}, 2022.

\bibitem{shlezinger2019viterbinet}
N.~Shlezinger, N.~Farsad, Y.~C. Eldar, and A.~J. Goldsmith, ``{ViterbiNet}: A
  deep learning based {Viterbi} algorithm for symbol detection,'' \emph{IEEE
  Transactions on Wireless Communications}, vol.~19, no.~5, pp. 3319--3331,
  2020.

\bibitem{shlezinger2021model2}
N.~Shlezinger, N.~Farsad, Y.~C. Eldar, and A.~J. Goldsmith, ``Model-based
  machine learning for communications,'' arXiv preprint arXiv:2101.04726, 2021.

\bibitem{shultzman2022}
A.~Shultzman, E.~Azar, M.~R.~D. Rodrigues, and Y.~C. Eldar, ``Generalization
  and estimation error bounds for model-based neural networks,'' in
  \emph{International Conference on Learning Representations}, 2023.

\bibitem{simoncelli_NaturalImageStatistics_2001}
E.~P. Simoncelli and B.~A. Olshausen, ``Natural image statistics and neural
  representation,'' \emph{Annual Review of Neuroscience}, vol.~24, no.~1, pp.
  1193--1216, 2001.

\bibitem{sitzmann_ImplicitNeuralRepresentations_2020a}
V.~Sitzmann, J.~Martel, A.~Bergman, D.~Lindell, and G.~Wetzstein, ``Implicit
  neural representations with periodic activation functions,'' in
  \emph{Conference on Neural Information Processing Systems}, 2020.

\bibitem{solomon2019deep}
O.~Solomon, R.~Cohen, Y.~Zhang, Y.~Yang, Q.~He, J.~Luo, R.~J. van Sloun, and
  Y.~C. Eldar, ``Deep unfolded robust {PCA} with application to clutter
  suppression in ultrasound,'' \emph{IEEE Transactions on Medical Imaging},
  vol.~39, no.~4, pp. 1051--1063, 2019.

\bibitem{soltanolkotabi_TheoreticalInsightsOptimization_2019}
M.~Soltanolkotabi, A.~Javanmard, and J.~D. Lee, ``Theoretical insights into the
  optimization landscape of over-parameterized shallow neural networks,''
  \emph{IEEE Transactions on Information Theory}, vol.~65, no.~2, pp. 742--769,
  2019.

\bibitem{sun2021scalable}
Y.~Sun, Z.~Wu, X.~Xu, B.~Wohlberg, and U.~S. Kamilov, ``Scalable plug-and-play
  {ADMM} with convergence guarantees,'' \emph{IEEE Transactions on
  Computational Imaging}, vol.~7, pp. 849--863, 2021.

\bibitem{szegedy2014intriguing}
C.~Szegedy, W.~Zaremba, I.~Sutskever, J.~Bruna, D.~Erhan, I.~Goodfellow, and
  R.~Fergus, ``Intriguing properties of neural networks,'' in
  \emph{International Conference on Learning Representations}, 2014.

\bibitem{tachella2022unsupervised}
J.~Tachella, D.~Chen, and M.~Davies, ``Unsupervised learning from incomplete
  measurements for inverse problems,'' in \emph{Conference on Neural
  Information Processing Systems}, 2022.

\bibitem{tancik_FourierFeaturesLet_2020a}
M.~Tancik, P.~Srinivasan, B.~Mildenhall, S.~{Fridovich-Keil}, N.~Raghavan,
  U.~Singhal, R.~Ramamoorthi, J.~Barron, and R.~Ng, ``Fourier features let
  networks learn high frequency functions in low dimensional domains,'' in
  \emph{Conference on Neural Information Processing Systems}, 2020.

\bibitem{tirer2018image}
T.~Tirer and R.~Giryes, ``Image restoration by iterative denoising and backward
  projections,'' \emph{IEEE Transactions on Image Processing}, vol.~28, no.~3,
  pp. 1220--1234, 2018.

\bibitem{ulyanov_DeepImagePrior_2018}
D.~Ulyanov, A.~Vedaldi, and V.~Lempitsky, ``Deep image prior,'' in
  \emph{Conference on Computer Vision and Pattern Recognition}, 2018.

\bibitem{vanveen_CompressedSensingDeep_2018}
D.~Van~Veen, A.~Jalal, E.~Price, S.~Vishwanath, and A.~G. Dimakis, ``Compressed
  sensing with deep image prior and learned regularization,''
  \emph{arXiv:1806.06438}, 2018.

\bibitem{venkatakrishnan2013plug}
S.~V. Venkatakrishnan, C.~A. Bouman, and B.~Wohlberg, ``Plug-and-play priors
  for model based reconstruction,'' in \emph{IEEE Global Conference on Signal
  and Information Processing}, 2013.

\bibitem{Vershynin2012}
R.~Vershynin, ``Introduction to the non-asymptotic analysis of random
  matrices,'' in \emph{Compressed Sensing: Theory and Applications}, Y.~Eldar
  and G.~Kutyniok, Eds.\hskip 1em plus 0.5em minus 0.4em\relax Cambridge
  University Press, 2012.

\bibitem{wang_PhaseImagingUntrained_2020}
F.~Wang, Y.~Bian, H.~Wang, M.~Lyu, G.~Pedrini, W.~Osten, G.~Barbastathis, and
  G.~Situ, ``Phase imaging with an untrained neural network,'' \emph{Light:
  Science \& Applications}, vol.~9, no.~1, pp. 1--7, 2020.

\bibitem{wei2019statistical}
X.~Wei, Z.~Yang, and Z.~Wang, ``On the statistical rate of nonlinear recovery
  in generative models with heavy-tailed data,'' in \emph{International
  Conference on Machine Learning}, 2019.

\bibitem{williams_DeepGeometricPrior_2019}
F.~Williams, T.~Schneider, C.~Silva, D.~Zorin, J.~Bruna, and D.~Panozzo, ``Deep
  geometric prior for surface reconstruction,'' in \emph{Conference on Computer
  Vision and Pattern Recognition}, 2019.

\bibitem{wu2020sparse}
K.~Wu, Y.~Guo, Z.~Li, and C.~Zhang, ``Sparse coding with gated learned
  {ISTA},'' in \emph{International Conference on Learning Representations},
  2020.

\bibitem{wu2019learning}
S.~Wu, A.~Dimakis, S.~Sanghavi, F.~Yu, D.~Holtmann-Rice, D.~Storcheus,
  A.~Rostamizadeh, and S.~Kumar, ``Learning a compressed sensing measurement
  matrix via gradient unrolling,'' in \emph{International Conference on Machine
  Learning}, 2019.

\bibitem{xiao2013proximal}
L.~Xiao and T.~Zhang, ``A proximal-gradient homotopy method for the sparse
  least-squares problem,'' \emph{SIAM Journal on Optimization}, vol.~23, no.~2,
  pp. 1062--1091, 2013.

\bibitem{xin2016maximal}
B.~Xin, Y.~Wang, W.~Gao, D.~Wipf, and B.~Wang, ``Maximal sparsity with deep
  networks?'' in \emph{Conference on Neural Information Processing Systems},
  2016.

\bibitem{yeh2016semantic}
R.~Yeh, C.~Chen, T.~Y. Lim, M.~Hasegawa-Johnson, and M.~N. Do, ``Semantic image
  inpainting with perceptual and contextual losses,'' \emph{arXiv preprint
  arXiv:1607.07539}, 2016.

\bibitem{yi2018outlier}
J.~Yi, A.~D. Le, T.~Wang, X.~Wu, and W.~Xu, ``Outlier detection using
  generative models with theoretical performance guarantees,'' \emph{arXiv
  preprint arXiv:1810.11335}, 2018.

\bibitem{yoo_TimeDependentDeepImage_2021}
J.~Yoo, K.~H. Jin, H.~Gupta, J.~Yerly, M.~Stuber, and M.~Unser,
  ``Time-dependent deep image prior for dynamic {MRI},'' \emph{IEEE
  Transactions on Medical Imaging}, vol.~40, no.~12, pp. 3337--3348, 2021.

\bibitem{zhang2022dive}
A.~Zhang, Z.~C. Lipton, M.~Li, and A.~J. Smola, ``Dive into deep learning,''
  2022, https://d2l.ai/.

\bibitem{zhang2019real}
L.~Zhang, G.~Wang, and G.~B. Giannakis, ``Real-time power system state
  estimation and forecasting via deep unrolled neural networks,'' \emph{IEEE
  Transactions on Signal Processing}, vol.~67, no.~15, pp. 4069--4077, 2019.

\bibitem{zhong_ReconstructingContinuousDistributions_2020}
E.~D. Zhong, T.~Bepler, J.~H. Davis, and B.~Berger, ``Reconstructing continuous
  distributions of {{3D}} protein structure from cryo-{{EM}} images,'' in
  \emph{International Conference on Learning Representations}, 2020.

\end{thebibliography}

\end{document}